\documentclass[sigconf]{aamas} 
\usepackage{balance} %
\usepackage{url}
\usepackage{hyperref}
\usepackage{changepage}
\usepackage[utf8]{inputenc} %
\usepackage[T1]{fontenc}    %
\usepackage{booktabs}       %
\usepackage{amsfonts}       %
\usepackage{nicefrac}       %
\usepackage{microtype}      %
\usepackage{multicol}
\usepackage{multirow}
\usepackage{mathtools}
\usepackage{verbatim}
\usepackage{caption}
\usepackage[ruled, noend, noline]{algorithm2e}
\usepackage{algorithmicx}
\usepackage[noend]{algpseudocode}
\usepackage{etoolbox}
\usepackage{subcaption}%
\usepackage{float}
\usepackage{wrapfig}
\usepackage{listings}
\usepackage{pifont}
\usepackage{enumitem}
\usepackage{tikz}
\usepackage{bbm}
\usepackage{tabularx}
\usepackage[export]{adjustbox}
\usepackage[noabbrev, capitalise]{cleveref}
\usepackage{colortbl}
\usepackage{utfsym}
\usepackage{amsmath, amsthm}
\usepackage{xcolor}
\usepackage{graphicx}
\usepackage{listings}

\setcopyright{ifaamas}
\acmConference[AAMAS '24]{Proc.\@ of the 23rd International Conference
on Autonomous Agents and Multiagent Systems (AAMAS 2024)}{May 6 -- 10, 2024}
{Auckland, New Zealand}{N.~Alechina, V.~Dignum, M.~Dastani, J.S.~Sichman (eds.)}
\copyrightyear{2024}
\acmYear{2024}
\acmDOI{}
\acmPrice{}
\acmISBN{}

\acmSubmissionID{234}

\title[MADRID]{Multi-Agent Diagnostics for Robustness via Illuminated Diversity}

\definecolor{lightgray}{rgb}{.9,.9,.9}
\definecolor{darkgray}{rgb}{.4,.4,.4}
\definecolor{purple}{rgb}{0.65, 0.12, 0.82}
\definecolor{darkgreen}{rgb}{0, 0.365, 0}

\definecolor{myblue}{HTML}{0000B5}
\definecolor{crimson}{HTML}{B30000}
\definecolor{mulberry}{rgb}{0.77, 0.29, 0.55}
\definecolor{palatinatepurple}{rgb}{0.41, 0.16, 0.38}
\definecolor{lt_red}{rgb}{1.0, 0.2, 0.4}
\definecolor{lt_blue}{rgb}{0.27, 0.6, 1}

\definecolor{ao(english)}{rgb}{0.0, 0.5, 0.0}

\makeatletter
\patchcmd{\@algocf@start}%
  {-1.5em}%
  {0pt}%
  {}{}%
\makeatother

\makeatletter
\patchcmd\algocf@Vline{\vrule}{\vrule \kern-0.4pt}{}{}
\patchcmd\algocf@Vsline{\vrule}{\vrule \kern-0.4pt}{}{}
\makeatother

\newcommand{\tocite}[1]{\textcolor{blue}{(To Cite et al, 2023)}}

        {\medskip}

\newcommand{\EO}{\mathop{\mathbb{E}}}

\newcommand{\PPOMDP}{\mathcal{M}}

\newcommand{\apply}[2]{#1_{#2}}

\author{Mikayel Samvelyan\footnotemark[1]}
\affiliation{
  \institution{UCL, Meta AI}
  }
  \country{}
\email{samvelyan@meta.com}

\author{Davide Paglieri\footnotemark[1]}
\affiliation{
  \institution{UCL}
    \country{}
}
\email{d.paglieri@cs.ucl.ac.uk}

\author{Minqi Jiang}
\affiliation{
  \institution{UCL, Meta AI}
    \country{}
}
\email{m.jiang@cs.ucl.ac.uk}

\author{Jack Parker-Holder}
\affiliation{
  \institution{UCL}
    \country{}
}
\email{j.parker-holder@ucl.ac.uk}

\author{Tim Rockt{\"a}schel}
\affiliation{
  \institution{UCL}
    \country{}
  }
\email{tim.rocktaschel@ucl.ac.uk}

\newcommand{\website}[0]{\url{https://sites.google.com/view/madrid-marl}}
\newcommand{\method}[0]{\textsc{MADRID}}
\newcommand{\methodlongemph}[0]{\emph{\textbf{M}ulti-\textbf{A}gent \textbf{D}iagnostics for \textbf{R}obustness via \textbf{I}lluminated \textbf{D}iversity}}
\newcommand{\methodlong}[0]{Multi-Agent Diagnostics for Robustness via Illuminated Diversity}

\begin{abstract}

\footnotetext[1]{Equal contribution} %

In the rapidly advancing field of multi-agent systems, ensuring robustness in unfamiliar and adversarial settings is crucial. 
Notwithstanding their outstanding performance in familiar environments, these systems often falter in new situations due to overfitting during the training phase.
This is especially pronounced in settings where both cooperative and competitive behaviours are present, encapsulating a dual nature of overfitting and generalisation challenges.
To address this issue, we present \emph{\methodlong{}} (\method{}), a novel approach for generating diverse adversarial scenarios that expose strategic vulnerabilities in pre-trained multi-agent policies. 
Leveraging the concepts from open-ended learning, \method{} navigates the vast space of adversarial settings, employing a target policy's regret to gauge the vulnerabilities of these settings. We evaluate the effectiveness of \method{} on the 11vs11 version of Google Research Football, one of the most complex environments for multi-agent reinforcement learning.
Specifically, we employ \method{} for generating a diverse array of adversarial settings for TiZero, the state-of-the-art approach which "masters" the game through 45 days of training on a large-scale distributed infrastructure. 
We expose key shortcomings in TiZero's tactical decision-making, underlining the crucial importance of rigorous evaluation in multi-agent systems.\footnote{Visuals are available at \website{}}%
\end{abstract}


\newcommand{\BibTeX}{\rm B\kern-.05em{\sc i\kern-.025em b}\kern-.08em\TeX}

\makeatletter
\gdef\@copyrightpermission{
	\begin{minipage}{0.3\columnwidth}
		\href{https://creativecommons.org/licenses/by/4.0/}{\includegraphics[width=0.90\textwidth]{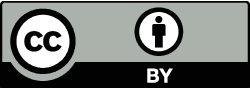}}
	\end{minipage}\hfill
	\begin{minipage}{0.7\columnwidth}
		\href{https://creativecommons.org/licenses/by/4.0/}{This work is licensed under a Creative Commons Attribution International 4.0 License.}
	\end{minipage}
	\vspace{5pt}
}
\makeatother

\begin{document}

\pagestyle{fancy}
\fancyhead{}

\maketitle 

\section{Introduction}

In recent times, multi-agent systems, particularly those designed to interact with humans, have emerged as a primary model for AI deployment in real-world scenarios~\cite{openai2023gpt4, anthropic_2023, badue2019selfdriving,touvron2023llama}. Although there have been significant successes in simulated environments, as evidenced by deep reinforcement learning (RL) in complex multi-agent games~\citep{alphago, muzero, alphastar, dota, wurman_outracing_2022}, the transfer from simulation to reality (sim2real) continues to pose challenges~\citep{hofer2021sim2real,Zhao2020SimtoRealTI}. 
Specifically, while these models demonstrate proficiency in known environments, they become highly susceptible to faulty behaviours in unfamiliar settings and adversarial situations~\citep{samvelyan2023maestro}.
Given their critical roles in human-centric applications, understanding and mitigating these susceptibilities becomes paramount for fostering more effective and reliable deployment of multi-agent AI systems in the future.

\begin{figure}
	\centering
	\includegraphics[width=\linewidth]{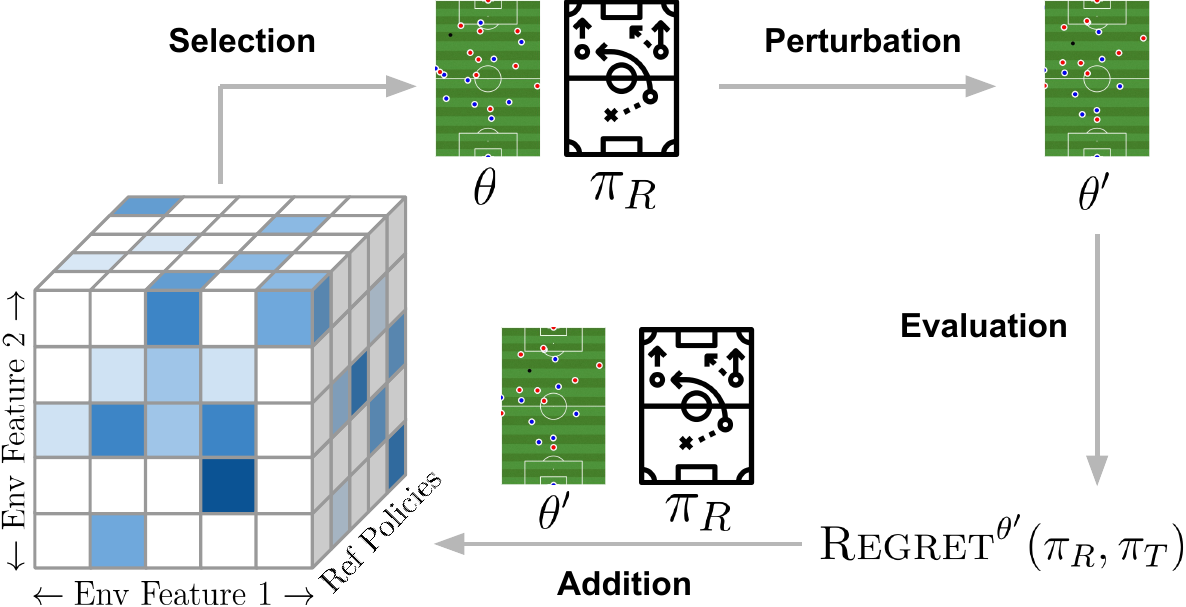}
	\caption{\label{fig:madrid}
		\textbf{Overview of \method{}}. Operating on a discretised grid with an added dimension for reference policies, \method{} archives environment variations (or levels) characterized by representative features, e.g., $(x,y)$ coordinates of the ball position in football. During each iteration, \method{} mutates a selected level, computes regret using its associated reference policy, and reincorporates levels with higher regret into the archive, effectively generating a diverse collection of adversarial levels.}
	\vspace{-0.4cm}
\end{figure}

The Achilles' heel of these multi-agent systems, contributing to their lack of robustness, is often their overfitting to the specific settings encountered during training~\citep{lanctot17unified}. 
This overfitting becomes notably evident in two-team zero-sum settings where both cooperative and competitive dynamics intertwine.
A primary manifestation of the overfitting between cooperative agents, especially when all agents in the group share the same set of network parameters (i.e., parameter sharing~\citep{foerster_learning_2016}), is in the agents becoming too accustomed to their training environments, leading to detailed coordination tailored to these specific conditions. As a consequence, when introduced to unfamiliar settings, their performance tends to falter. Concurrently, there is also an overfitting to specific opponent teams they have trained against. Instead of developing a flexible strategy that can withstand a variety of opponents, their strategies might be overly optimised to counteract the strategies of familiar adversaries. 
These dual forms of overfitting—both to the environment and to opponents—render such settings as perfect platforms to probe for vulnerabilities~\cite{tuys2021game}. Furthermore, it is crucial to pinpoint a diverse set of adversarial scenarios for a holistic diagnostic of robustness, shedding light on possible shortcomings from various perspectives.%

Given these challenges, we introduce \methodlongemph{} (\method{}), a novel method for systematically generating a diverse collection of adversarial settings where pre-trained multi-agent policies make strategic mistakes. 
To this end, \method{} employs approaches from quality-diversity (QD)~\citep{lehman2011abandoning, Cully2018Quality}, a family of evolutionary algorithms that aim to generate a large collection of high-performing solutions each with their own unique characteristics.

\method{} incorporates MAP-Elites~\citep{mouret2015illuminating}, a simple and effective QD approach, to systematically explore the vast space of adversarial settings. By discretising the search space, \method{} iteratively performs selection, mutation, and evaluation steps, continually refining and expanding the repertoire of high-performing adversarial scenarios within its archive (see \cref{fig:madrid}). 
A crucial feature of \method{} is its employment of the target policy's \textit{regret}— the gap in performance between the optimal and target policy—to quantify the quality of adversarial settings. 
Regret is shown to be an effective metric for identifying situations where RL agents underperform in both single-agent \citep{paired, jiang2021robustplr, parker-holder2022evolving, mediratta2023stabilizing} and multi-agent \citep{samvelyan2023maestro} domains.
\method{} estimates a lower bound on the true regret by utilising a collection of reference policies~\cite{alphastar,garnelo2021pick}, which are not necessarily required to be high-performing. \method{} identifies situations where these reference policies surpass the target one, thereby providing a clear illustration of superior performance in given situations.

To evaluate \method{}, we concentrate specifically on one of the most challenging multi-agent domains, namely the fully decentralised 11 vs 11 variation of Google Research Football \citep[GRF,][]{kurach2020google}.
This simulated environment is based on the popular real-world sport of football (a.k.a. soccer) and requires two teams of agents to combine short-term control techniques with coordinated long-term global strategies.
GRF represents a unique combination of characteristics not present in other RL environments~\citep{tizero}, namely multi-agent cooperation (within each team), competition (between the two teams), sparse rewards, large action and observation spaces, and stochastic dynamics.
While many of the individual challenges in GRF, including multi-agent coordination~\citep{rashid2018qmix,yu2022the}, long-term planning~\citep{Ecoffet2020FirstRT} and non-transitivity~\citep{balduzzi19open-ended,czarnecki2020real}, have been studied extensively in isolation, learning highly-competitive GRF policies has long remained outside the reach of RL methods.
TiZero~\citep{tizero}, a recent multi-agent RL approach, learned to "master" the fully decentralised variation of GRF from scratch for the first time, using a hand-crafted curriculum, reward shaping, and self-play. Experimentally, TiZero has shown impressive results and outperformed previous methods by a large margin after an expensive training lasting 45 days on a large-scale distributed training infrastructure. %

We apply \method{} on GRF by targeting TiZero to diagnose a broad set of scenarios in which it commits tactical mistakes. Our extensive evaluations reveal diverse settings where TiZero exhibits poor performance, where weaker policies can outperform it. Specifically, \method{} discovers instances where TiZero is ineffective near the opponent's goal, demonstrates a marked inability to comprehend the offside rule effectively, and even encounters situations of scoring accidental own goals. 
These findings highlight the latent vulnerabilities within even highly trained models and demonstrate that there is much room for improving their robustness.
Our analysis showcases the value of identifying such adversarial settings in offering new insights into the hidden weaknesses of pretrained policies that may otherwise appear highly robust.

\section{Background}\label{sec:Background}

\paragraph{\textbf{Underspecified Stochastic Games}}
In this work, we consider \textit{Underspecified Stochastic Games (USG)}, i.e., stochastic games \citep{shapley1953stochastic} with underspecified parameters of the environment.
A USG for an environment with $n$ agents is defined by a set of states $\mathcal{S}$ and sets of per-agent actions $\mathcal{A}_1,...,\mathcal{A}_n$ and observations $\mathcal{O}_1,...,\mathcal{O}_n$.
Each agent $i$ select actions using a stochastic policy $\pi_i: \mathcal{O}_i \times \mathcal{A}_i \mapsto [0,1]$.
$\Theta$ defines the set of \textit{free parameters} of the environment that is incorporated into the transition function $\mathcal{T} : \mathcal{S} \times \Theta \times \mathcal{A}_1 \times ... \times \mathcal{A}_n \mapsto \mathcal{S}$ producing the next state based on the actions of all agents.
Each agent $i$ receives observations $\mathbf{o}_i : \mathcal{S} \mapsto \mathcal{O}_i$ correlated with the current state and reward $r_i : \mathcal{S} \times \mathcal{A}_i \mapsto \mathbb{R}$  as a function of the state and agent’s action.
The goal of each agent $i$ is to maximise its own total expected return $R_i = \sum_{t=0}^T \gamma^t r^t_i$ for the time horizon $T$, where $\gamma$ is a discount factor.

Each configuration of the free parameter $\theta \in \Theta$, which is often called a \emph{level} \citep{jiang2021robustplr, parker-holder2022evolving}, defines a specific instantiation of the environment $\apply{\PPOMDP}{\theta}$. %
For example, this can correspond to different positions of the walls in a maze, or locations of players and the ball in a football game.
USG is a multi-agent variation of Underspecified POMDPs~\citep{paired} and the fully observable variant of UPOSGs~\citep{samvelyan2023maestro}.

\paragraph{\textbf{Quality-Diversity}} Quality-diversity (QD) is a family of methods used to find a \textit{diverse collection} of solutions that are \textit{performant} and span a meaningful spectrum of solution characteristics~\citep{lehman2011abandoning, Cully2018Quality}. %
The performance of solution $x\in \mathcal{X}$ is measure using the \textit{fitness} 
$:\mathcal{X} \mapsto \mathbb{R}$ function.
The diversity of solutions is typically measured using the \emph{feature\_descriptor}  $:\mathcal{X} \mapsto \mathbb{B}$ function that maps a solution into the feature space $\mathbb{B}=\mathbb{R}^K$ that describes specific characteristics of the solution, such as behavioural properties or visual appearance.

\begin{algorithm}[]
	\SetAlgoLined
	\caption{MAP-Elites~\citep{mouret2015illuminating}}
	\label{alg:map_elites}
	\textbf{Initialise:} Empty $N$-dimensional grids for solutions $X$ and performances $\mathcal{P}$\\
	Populate $n$ cells of $X$ with random solutions and corresponding cells of $P$ with their fitness scores \\
	\For{$i = \{1,2, \dots\}$} {
		Sample solution $x$ from $X$ \\
		Get solution $x'$ from $x$ via random mutation \\
		$p' \leftarrow fitness(x')$ \\
		$b' \leftarrow feature\_descriptor(x')$ \\
		\If{$\mathcal{P}(b')=\emptyset~or~\mathcal{P}(b') < p'$} {
			$\mathcal{P}(b') \leftarrow p'$\\
			$X(b') \leftarrow b'$\\
		}
	}
\end{algorithm}
\DecMargin{1em}

\paragraph{\textbf{MAP-Elites}}
\textit{MAP-Elites} is a simple and effective QD method \citep{mouret2015illuminating}.
Here, the descriptor space $\mathbb{B}$ is discretised and represented as an initially empty $N \leq K$ dimensional grid, also referred to as the \emph{archive}.
The algorithm starts by generating an arbitrary collection of candidate solutions. %
At each iteration, a solution is randomly selected among those in the grid.
A new solution is obtained by mutating the selected solution, which is then evaluated
and mapped to a cell of the grid based on its feature descriptor. 
The solution is then placed in the corresponding cell of the grid if it has a higher fitness than the current occupant, or if the cell is empty.
This cycle of selection, mutation, and evaluation is repeated, progressively enhancing both the diversity (coverage) and the quality (fitness) of the collection.
The pseudo-code of MAP-Elites is presented in \cref{alg:map_elites}.

\section{MADRID}\label{sec:Method}

In this section, we describe \methodlongemph{} (\method{}), a novel method for automatically generating diverse adversarial settings for a \textit{target} pre-trained policy $\pi_T$.
These are settings that either deceive the policy, forcing it to produce unintended behaviour, or where the policy inherently performs poorly, deviating from the optimal behaviour.
For USGs, these settings correspond to particular environment levels $\theta \in \Theta$ that have been procedurally generated. %

For quantifying adversarial levels, we make use of the target policy's \textit{regret} in level $\theta$, i.e., the difference in utility between the optimal policy $\pi^*$ and $\pi_T$ :
$$
\textsc{Regret}^\theta(\pi^*,\pi_T) = V^\theta(\pi^*,\pi_T) - V^\theta(\pi_T,\pi_T),
$$
where $V_\theta(\pi_A, \pi_B) = \EO[\sum_{t=0}^{T} \gamma^tr_t^{A}]$ is the value of a policy $\pi_A$ against policy $\pi_B$ in $\theta$.\footnote{Note that here, for the simplicity of the notation, we assume a two-team zero-sum setting. $\pi_A$ and $\pi_B$ describe the policies for groups of agents, either through a centralised controller or decentralised policies that employ parameter sharing. However, \method{} can be applied for more general multi-agent settings.}

Regret is a suitable metric for evaluating adversarial examples in pre-trained models. It provides a measure that directly quantifies the suboptimality of a model's decisions. While a high regret value serves as a glaring indicator of how far off a model's behaviour is from the optimal choice, a low regret indicates the model's decisions are closely aligned with the optimal choice. The importance of regret becomes even more pronounced when considering the varied scenarios in which a model might be deployed. Therefore, by investigating regret across \textit{diverse} situations, we can not only pinpoint specific vulnerabilities of a model but also ensure robustness in previously unseen scenarios.

Since the optimal policy is usually unavailable, \method{} relies on utilising a collection of \textit{suboptimal} policies $\Pi_{R} = \bigcup_{i=1}^{M}\pi_{i} $ for estimating the lower bound on true regret. 
Specifically, the goal is to find adversarial levels that maximise the gap in utility acquired through a \textit{reference} policy $\pi_{i} \in \Pi_{R}$ and target policy $\pi_T$. %
Utilising a collection of diverse reference policies can be advantageous in the absence of a true optimal policy, since each of these reference policies may excel in a unique set of levels~\cite{samvelyan2023maestro}.

\begin{algorithm}
	\SetAlgoLined
	\caption{\method{}}
	\label{alg:madrid}
	\textbf{Input:} Target policy $\pi_T$, a collection of reference policies $\Pi_R$, $level\_descriptor: \Theta \mapsto \mathbb{R}^N$ function\\
	\textcolor{gray}{\textit{\# Initialise a discretised grid, with an added dimension for $\Pi_R$, to archive levels and regret scores.}}\\
	\textbf{Initialise:} Empty $N+1$-dimensional grids for levels $X$ and regret estimates $\mathcal{P}$\\
	Populate $n$ cells of $X$ with randomly generated levels and corresponding estimated regret in $\mathcal{P}$ \\
	\For{$i = \{1,2, \dots\}$} {
		\textcolor{gray}{\textit{\# Sample a level $\theta$ and corresponding reference policy $\pi_R$ from $X$.}}\\
		$\theta, \pi_R \sim X$ \\
		\textcolor{gray}{\textit{\# Perform level mutation by adding Gaussian noise.}}\\
		$\theta' \leftarrow \theta + \mathcal{N}(0,\,\sigma^{2})$ \\
		\textcolor{gray}{\textit{\# Estimate the regret of $\pi_T$ on $\theta'$ using $\pi_R$.}}\\
		$\widetilde{r}' \leftarrow V^{\theta'}(\pi_R, \pi_T) - V^{\theta'}(\pi_T, \pi_T)$ \\
		$b' \leftarrow level\_descriptor(\theta')$ \\
		\If{$\mathcal{P}(b',\pi_R)=\emptyset~or~\mathcal{P}(b',\pi_R) < \widetilde{r}'$} {
			$\mathcal{P}(b',\pi_R) \leftarrow \widetilde{r}'$\\
			$X(b',\pi_R) \leftarrow b'$\\
		}
	}
\end{algorithm}

\method{} casts the task of generating a diverse array of adversarial levels for each reference policy as a QD search problem.
Specifically, \method{} uses MAP-Elites to systematically generate levels from $\Theta$ by discretising the feature space of levels into an $N$-dimensional grid, with an additional dimension representing the corresponding reference policy from $\Pi_R$.
Using a discretised grid of MAP-Elites provides interpretability to the adversarial examples found in \method{} given that each cell defines specific environment parameters, alongside a reference policy which outperforms the target under these parameters.

\method{} starts by populating the grid with randomly generated initial levels for each reference policy. During the iterative process, levels are selected from the grid to undergo mutation, followed by regret estimation. Each mutated level is then mapped to a specific cell in the grid based on its features and replaces the existing occupant if the mutated level has higher regret or the corresponding cell is unoccupied. This procedure ensures a thorough exploration and exploitation of the environment design space, allowing \method{} to generate levels that are both diverse and have high regret. \cref{fig:madrid} illustrates this process. \cref{alg:madrid} provides the pseudocode of the method.

\section{Experimental Setting}\label{sec:experiment}

\begin{figure*}
	\centering
	\begin{minipage}{.7\textwidth}
		\centering
		\includegraphics[height=5.6cm]{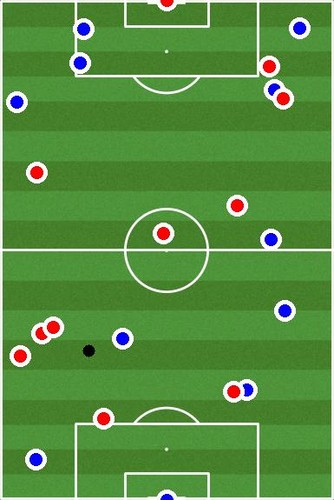}
		\includegraphics[height=5.6cm]{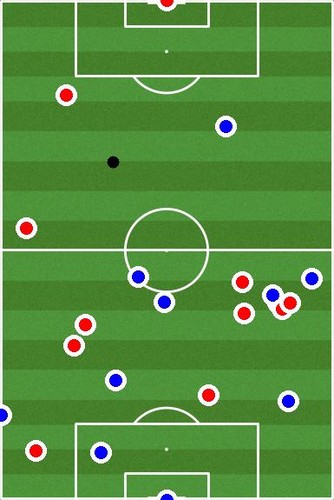}
		\includegraphics[height=5.6cm]{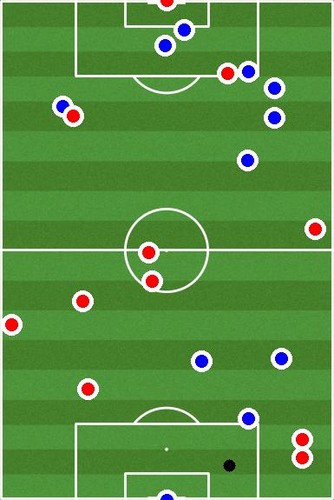}
		\caption{\label{fig:random_levels}Examples of randomly generated levels on Google Research Football.~\vspace{.35cm}}
	\end{minipage}%
	\begin{minipage}{.29\textwidth}
		\centering
		\includegraphics[height=5.6cm]{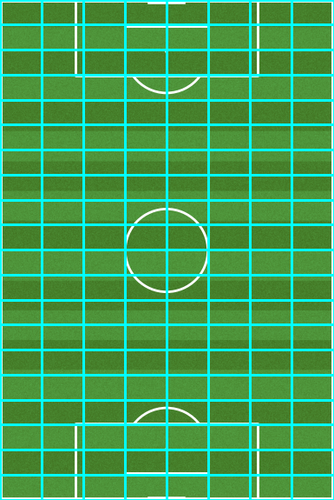} %
		\caption{Dividing the field in 160 grids using the ball $(x,y)$ coordinates.}
		\label{fig:160_grids}
	\end{minipage}
\end{figure*}

We evaluate \method{} on Google Research Football domain~\citep[GRF,][]{kurach2020google}.
Our experiments seek to (1) showcase the effectiveness of \method{} in generating diverse adversarial settings for a target state-of-the-art pre-trained RL model, (2) analyse the adversarial settings generated by \method{} to find key weaknesses of the target model, (3) validate the design choices of \method{} by comparing it to two ablated baselines.
Given its strong performance and usage in related works, Covariance Matrix Adaptation MAP-Elites \citep[CMA-ME,][]{fontaine2020coveriance} serves as the base MAP-Elites method in our experiments.
We provide full environment descriptions in \cref{appendix:env} and implementation details in \cref{appendix:implementation}.

\paragraph{\textbf{Baselines}} We compare \method{} against two baselines:
The \textit{targeted baseline} uses a MAP-Elites archive but randomly samples levels from scratch, rather than evolving previously discovered high-regret levels from the grid. Consequently, it does not leverage the discovered stepping stones throughout the search process~\citep{lehman2011abandoning}.
The \textit{random baseline} samples levels randomly from scratch without maintaining an archive of high-regret levels. %

\subsection*{Environment}\label{sec:env}

We use \method{} to find adversarial scenarios for TiZero, the state-of-the-art model for GRF. TiZero was trained via a complex regime on large-scale distributed infrastructure \citep{tizero} over 45 days. In particular, we aim to generate adversarial levels whereby the decentralised agents in TiZero make a diverse array of strategic errors, as highlighted by better behaviours of the reference policy.

GRF is a complex open-source RL environment designed for training and evaluating agents to master the intricate dynamics of football, one of the world’s most celebrated sports.
It offers a physics-based 3D simulation that tasks the RL policy with controlling a team of players to penetrate the opponent's defence while passing the ball among teammates to score goals. 
GRF is a two-team zero-sum environment that has long been considered one of the most complex multi-agent RL benchmarks due to a unique combination of challenges~\citep{huang2021tikick,wen2022multi,tizero}, such as multi-agent cooperation, multi-agent competition, sparse rewards, large action and observation spaces, and stochastic dynamics.\footnote{Highlighting the stochasticity of the GRF environment, a shot from the top of the box can lead to various outcomes, underscoring that not every action results in a predictable outcome.}

In this work, we focus on the fully decentralised 11 vs 11 version of the environment where each of the $10$ RL agents on both sides controls an individual player on the field.\footnote{The goalkeepers are controlled by the game AI.} Following \citep{tizero}, each agent receives a $268$-dimensional feature vector as an observation including the player's own characteristics, information about the ball, players of the both sides, as well as general match details.
The action space of agents consists of $19$ discrete actions, such as moving in 8 directions, sprinting, passing, and shooting.

To apply \method{} on GRF, we utilise procedurally generated levels each represented as a vector consisting of $(x,y)$ coordinates of $20$ players\footnote{The goalkeepers' position positions are always near their own goals.} and the ball. 
The position of the ball on the field serves as a convenient descriptor for levels in GRF because it accommodates diverse scenarios, ranging from attacking to defending on either side of the field. Therefore, we use the $x$ and $y$ coordinates of the ball as the two environment features in \method{}'s archive of levels.
This leads to a categorisation of levels into $160$ uniformly spaced cells across the football field, as illustrated in \cref{fig:160_grids}. Given that we are interested in evaluating TiZero in specific adversarial levels, in our experiments we restrict the episode length to $128$ steps taking place at the beginning of the game.

The third axis for the MAP-Elites archive indexes the reference policies $\Pi_R$. 
In our experiments, we make use of $48$ checkpoints of TiZero saved throughout its training \citep{tizero}, as well as three built-in bots in GRF with varying difficulties (easy, medium, and hard). 
For each reference policy, we initialise the grid with randomly sampled levels that assign random locations to players and the ball. \cref{fig:random_levels} illustrates some of the randomly generated levels. 

At each iteration of \method{}, we sample a level and reference policy pair $(\theta, \pi_R)$. The level is then mutated by adding Gaussian noise to the $(x, y)$ positions of the players and the ball in the field.
The fitness of each solution is estimated by computing TiZero's regret, which is the difference in performance between the selected reference policy $\pi_R$
and TiZero's policy $\pi_T$. In both cases, we estimate the regret against the TiZero policy on the level $\theta$ as:
\begin{equation}
	\widetilde{Regret}(\theta, \pi_T, \pi_R) = V^\theta(\pi_R, \pi_T) - V^\theta(\pi_T, \pi_T),
	\label{eq:regret_estimate}
\end{equation}
which corresponds to the difference in cross-play and self-play values between the reference and target policies. The performance on a given level $\theta$ between two policies $\pi_A$ and $\pi_B$ is the reward for scoring a goal:
\begin{equation}
	\label{eq:value}
	V^\theta(\pi_A, \pi_B) = 
	\begin{cases}
		1 & \text{if }\pi_A\text{ scores} \\
		0 & \text{if no goal is scored} \\
		-1 & \text{if }\pi_B\text{ scores}
	\end{cases}
\end{equation}
Upon scoring a goal by either of the sides, the level terminates. Given the non-deterministic nature of GRF, we account for variability by calculating the average regret across $4$ repetitions of the same pair of level $\theta$ and reference policy $\pi_R$.

\begin{figure*}
	\centering
	\begin{subfigure}[b]{0.32\textwidth}
		\centering
		\includegraphics[height=4.7cm]{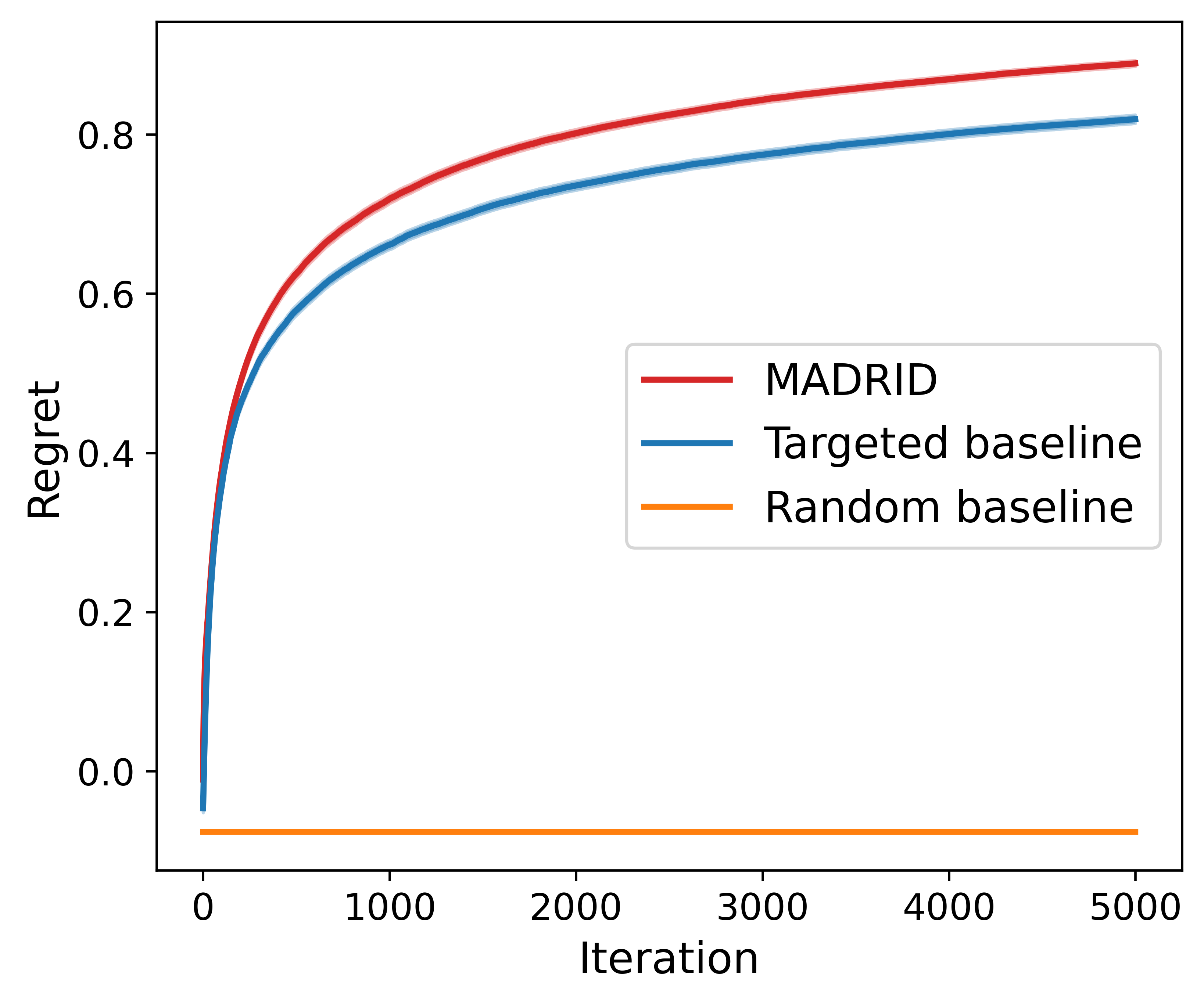}
		\caption{Estimated regret at each iteration.}
		\label{fig:average_regret_steps}
	\end{subfigure}
	\begin{subfigure}[b]{0.66\textwidth}
		\centering
		\includegraphics[height=4.7cm]{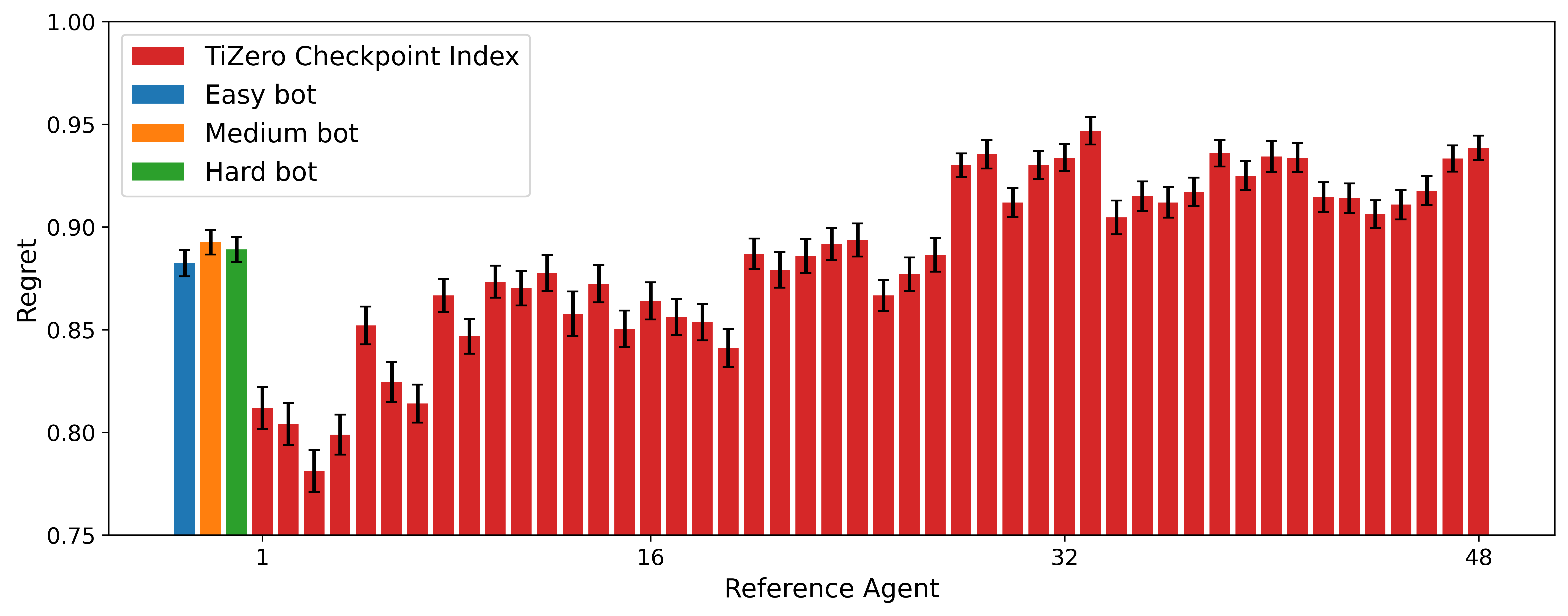}
		\caption{Final estimated regret of TiZero over reference policies using \method{}.} %
		\label{fig:grf_results_final_regret}
	\end{subfigure}
	\begin{subfigure}[b]{0.32\textwidth}
		\centering
		\includegraphics[height=4.7cm]{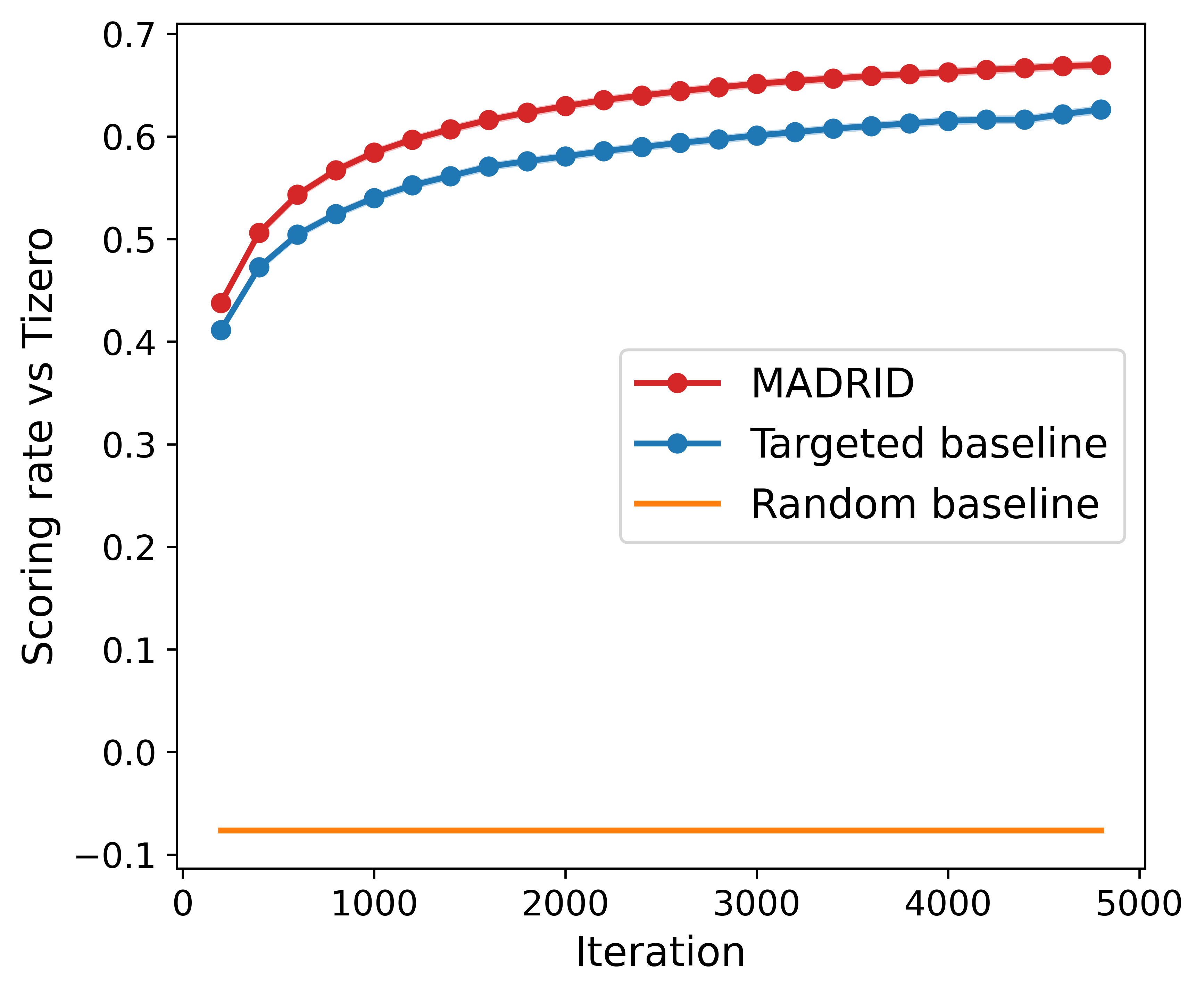}
		\caption{Scoring rate vs TiZero at each iteration.}
		\label{fig:average_scoring_steps}
	\end{subfigure}
	\begin{subfigure}[b]{0.66\textwidth}
		\centering
		\includegraphics[height=4.7cm]{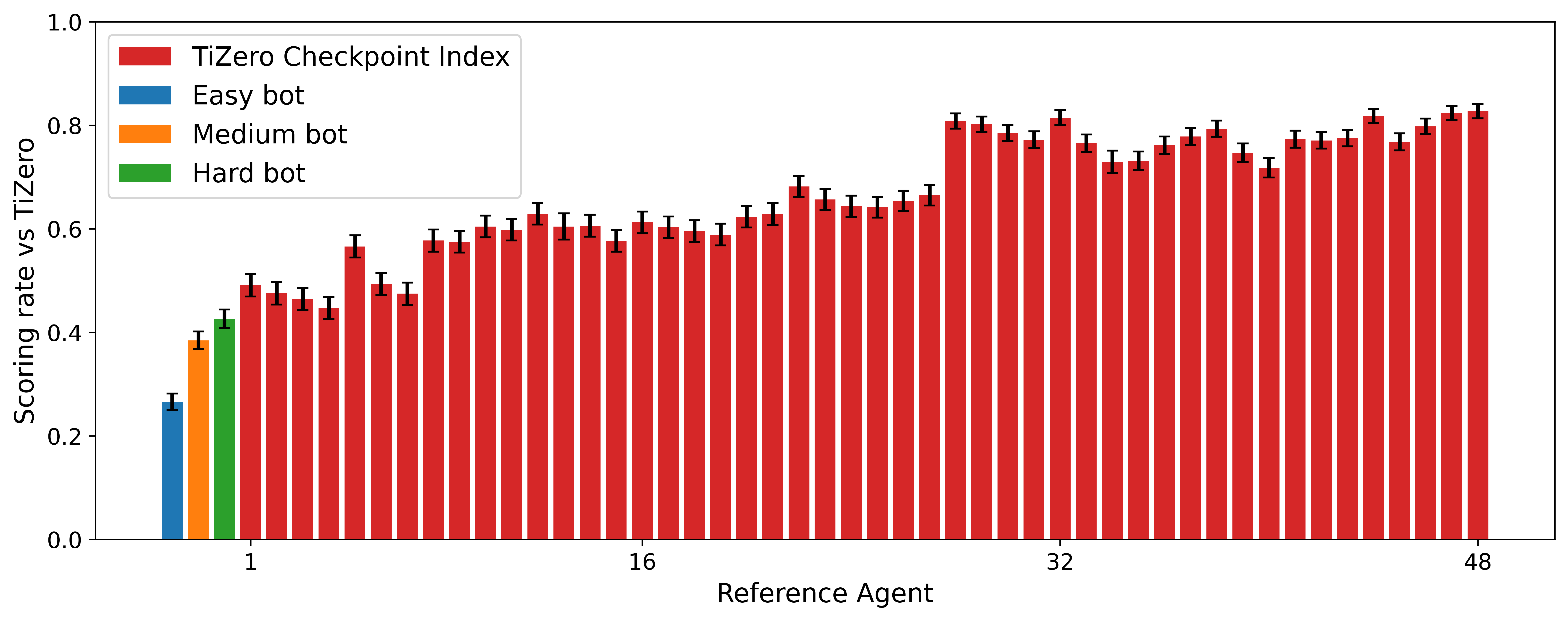}
		\caption{Final estimated scoring rate vs TiZero over reference policies using \method{}.} %
		\label{fig:grf_results_final_scoring}
	\end{subfigure}
	\vspace{-0.2cm}
	\caption{The estimated regret and goal score rate against TiZero in GRF are illustrated through each iteration for 51 reference agents, as shown in (a) and (c). The final values are presented in figures (b) and (d). Standard error over 3 random seeds is shown.}

\end{figure*}

\section{Results and Discussion}\label{sec:results}

In our analysis of targeting TiZero on GRF, we closely examine the performance of \method{} and baselines.
\cref{fig:average_regret_steps} displays the average estimated regret values for all $160$ cells within the MAP-Elites archive across the entire collection of reference policies. Here, \method{} outperforms both baselines. The \textit{random baseline} exhibits a negative value close to $0$, as TiZero proves to be a stronger policy than all the reference policies on entirely random game levels. On the other hand, the \textit{targeted baseline} iteratively populates the archive with high-regret levels, closely resembling \method{}'s performance at the early stages of iterations. However, as the iterations continue, it lags behind due to its failure to capitalise on previously identified high-regret levels that serve as stepping stones for the next iterations.

In \cref{fig:grf_results_final_regret}, we illustrate the variation in estimated final regret scores across the reference policies. Here, the regret increases as we move to higher-ranked agents. 
The heuristic bots display regret levels that are on par with the intermediate checkpoints of TiZero.

As we approximate the regret using the difference between cross-play (XP) and self-play (SP) between reference and TiZero policies (see Equations \ref{eq:regret_estimate} and \ref{eq:value}), a regret estimate of $1$ for an adversarial level $\theta$ can be achieved in two situations. First, the reference policy scores against TiZero in XP, while TiZero cannot score in SP. Second, TiZero concedes a goal in SP in $\theta$. 
Intriguingly, our findings reveal that for around 90\% of the adversarial levels generated by \method{}, a nominally weaker reference policy outperforms TiZero. This emphasises \method{}'s capability in exposing adversarial levels where even state-of-the-art policies are prone to missteps.

\begin{figure}[h!]
	\centering
	\includegraphics[height=4.2cm]{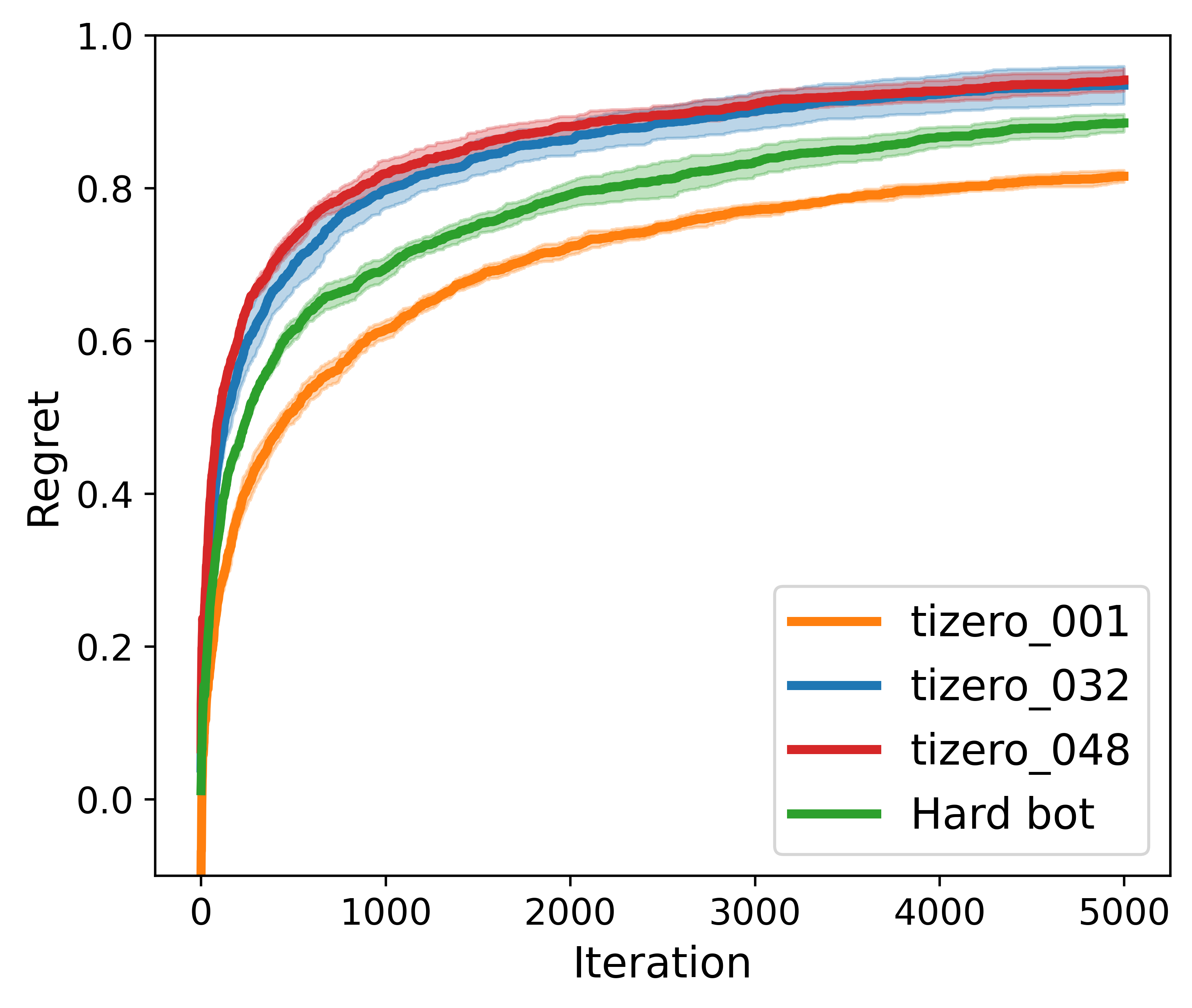}
	\vspace{-0.4cm}
	\caption{\method{}'s estimated regret over different reference policies after each iteration on GRF (mean and standard error over 3 seeds).}
	\label{fig:average_regret_individual_lines}
\end{figure}

\begin{figure*}
	\begin{subfigure}[b]{0.24\textwidth}
		\centering
		\includegraphics[width=\textwidth]{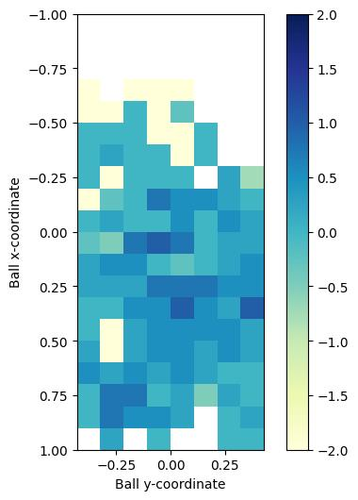}
		\caption{25 iterations.}
	\end{subfigure}
	\hfill
	\begin{subfigure}[b]{0.24\textwidth}
		\centering
		\includegraphics[width=\textwidth]{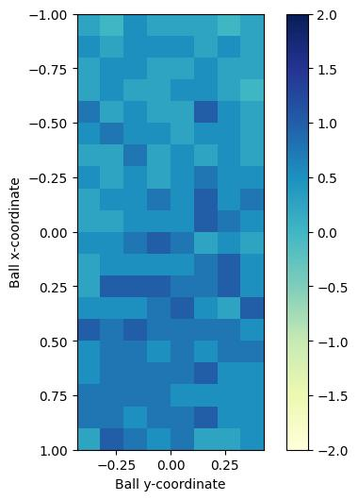}
		\caption{200 iterations.}
	\end{subfigure}
	\hfill
	\begin{subfigure}[b]{0.24\textwidth}
		\centering
		\includegraphics[width=\textwidth]{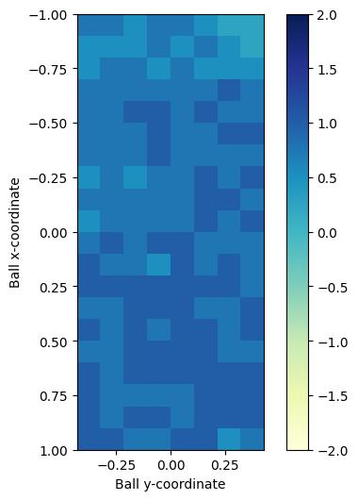}
		\caption{1000 iterations.}
		\label{fig:pipeline_llm}
	\end{subfigure}
	\hfill
	\begin{subfigure}[b]{0.24\textwidth}
		\centering
		\includegraphics[width=\textwidth]{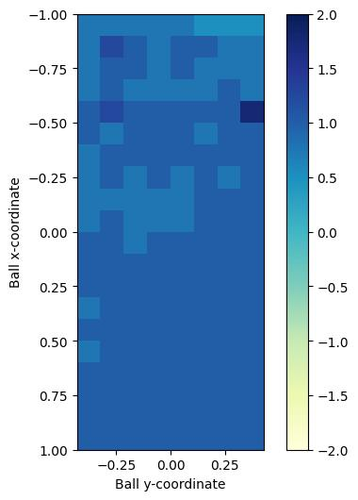}
		\caption{5000 iterations.}
		\label{fig:pipeline_llm_final}
	\end{subfigure}
	\caption{The estimated regret in \method{}'s archive at various iterations with respect to TiZero-048 reference policy.}
	\label{fig:football_archives}
\end{figure*}

\cref{fig:average_scoring_steps} and \cref{fig:grf_results_final_scoring} illustrate the estimated rate of goals scored against TiZero by the reference policies on adversarial levels produced by \method{} and baselines. We can see that approximately 70\% of the time across all reference policies, the reference policy scored a goal against TiZero in a short period of time.\footnote{The levels last only $128$ environment steps, which is a short episode compared to the $3000$ steps for the full game.} It should be noted that within the remaining 30\%, the majority of instances result in no scored goals. %

\cref{fig:average_regret_individual_lines} highlights the difference in performance for selected reference policies. Notably, the higher-ranked checkpoints of TiZero, saved at the later stages of its training, can be used to identify more severe vulnerabilities, as measured using the regret estimate.

\cref{fig:football_archives} shows the evolution of \method{}'s archive for a specific reference policy, illustrating its search process over time. Initially, the grid is sparsely filled with low-regret levels. However, as iterations progress, \method{} generates high-regret levels that progressively populate the entire grid. This shows that \method{} can discover high-regret levels anywhere on the football field. On average, we notice that higher-level scenarios tend to be located towards the positive $x$ coordinates. These correspond to situations where the ball is close to the opponent's goal from the perspective of the reference policy. While most regret scores tend to have uniform values around similar positions on the field, in \cref{fig:pipeline_llm_final} the grid also includes an adversarial level with an estimated regret of $1.75$. This indicates that \method{} found a level where the reference policy scores against TiZero in XP, while TiZero concedes a goal in SP.

\subsection{\textbf{Qualitative Analysis}}\label{sec:football_analysis}
We conduct a qualitative analysis of the adversarial levels identified by \method{} on GRF by visualising the highest ranking levels in the archive across all reference policies. We provide a selection of these examples below, with a comprehensive list available in \cref{appendix:football}. Full videos of all identified vulnerabilities can be found at \website{}.

\begin{figure*}
	\begin{subfigure}[b]{0.329\textwidth}
		\centering
		\includegraphics[width=\textwidth]{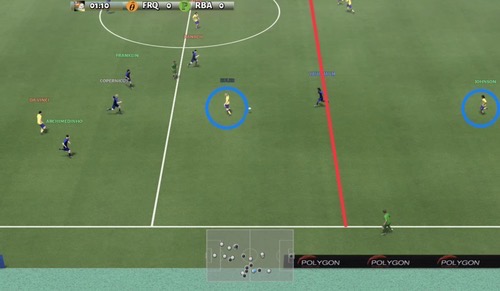}
		\caption{Initial player and ball positions in the level. TiZero is about to pass the ball to a teammate.~\vspace{.35cm}}
	\end{subfigure}
	\hfill
	\begin{subfigure}[b]{0.329\textwidth}
		\centering
		\includegraphics[width=\textwidth]{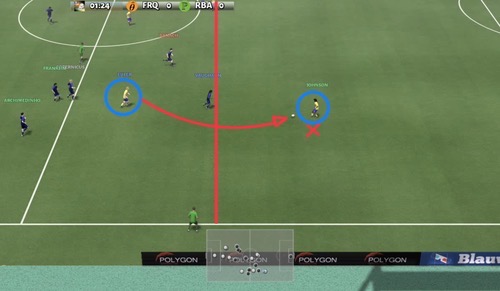}
		\caption{The receiving player is clearly offside, thus a freekick is awarded to the opponent team.~\vspace{.35cm}}
	\end{subfigure}
	\hfill
	\begin{subfigure}[b]{0.329\textwidth}
		\centering
		\includegraphics[width=\textwidth]{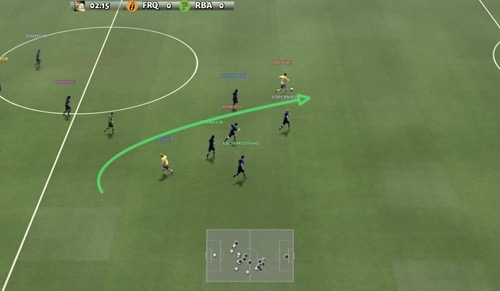}
		\caption{Reference policy does not pass to the offside player and directly runs towards the goal to score.}
	\end{subfigure}
	\caption{Adversarial example of offsides.}
	\label{fig:offside}
\end{figure*}

\paragraph{\textbf{Offsides}} Despite its strong performance under standard evaluations, TiZero frequently falls victim to erroneously passing the ball to players unmistakably in offside positions, as shown in \cref{fig:offside}
This observation highlights TiZero's lack of a deep understanding of the rules of the game. In contrast, the reference policies abstain from passing the ball to offside players, resulting in successful scoring outcomes.\footnote{Player are offside when they are in the opponent's half and any part of their body is closer to the opponent's goal line than both the ball and the second-last opponent. Usually one of the two opponents is the goalkeeper. When the ball is passed to a player who is offside, a free kick is awarded to the opponent's team.}

\begin{figure*}
	\begin{subfigure}[b]{0.329\textwidth}
		\centering
		\includegraphics[width=\textwidth]{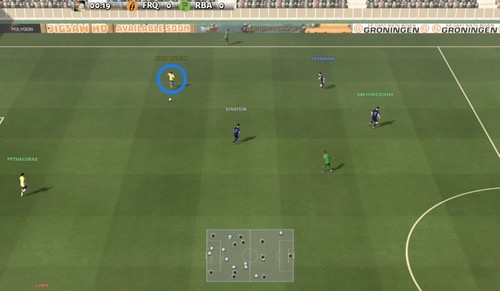}
	\end{subfigure}
	\hfill
	\begin{subfigure}[b]{0.329\textwidth}
		\centering
		\includegraphics[width=\textwidth]{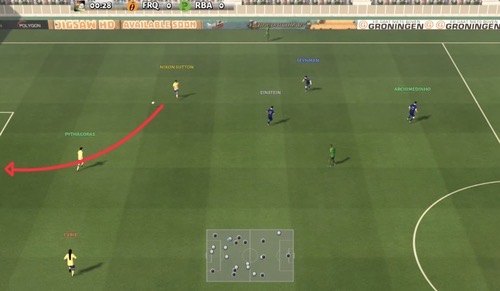}
	\end{subfigure}
	\hfill
	\begin{subfigure}[b]{0.329\textwidth}
		\centering
		\includegraphics[width=\textwidth]{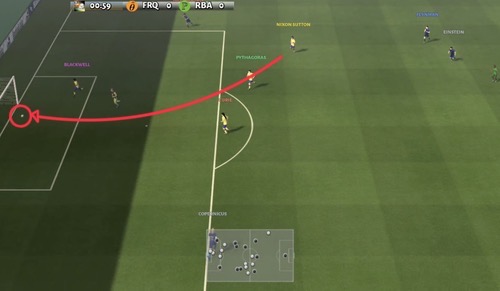}
	\end{subfigure}
	\caption{Adversarial example of an own goal. TiZero gets tricked and shoots in its own goal.}
	\label{fig:owngoal}
\end{figure*}

\begin{figure*}
	\begin{subfigure}[b]{0.329\textwidth}
		\centering
		\includegraphics[width=\textwidth]{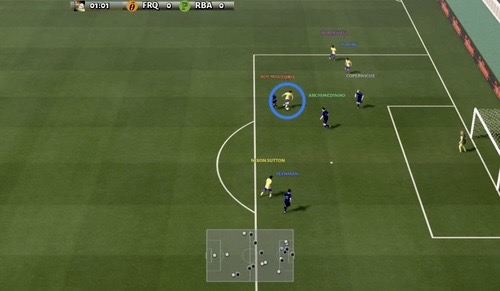}
	\end{subfigure}
	\hfill
	\begin{subfigure}[b]{0.329\textwidth}
		\centering
		\includegraphics[width=\textwidth]{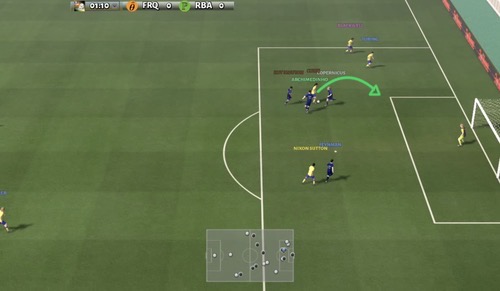}
	\end{subfigure}
	\hfill
	\begin{subfigure}[b]{0.329\textwidth}
		\centering
		\includegraphics[width=\textwidth]{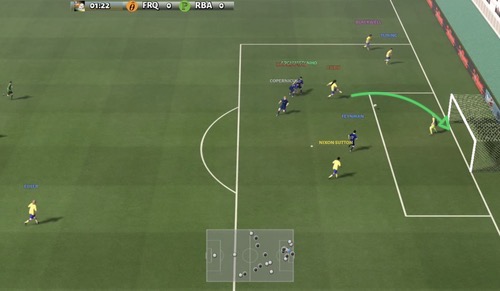}
	\end{subfigure}
	\caption{Adversarial example of a slow-running opponent. Three TiZero-controlled defenders are not able to stop a simple slow-running opponent, who walks past them and scores.}
	\label{fig:slow_running}
\end{figure*}

\begin{figure*}
	\begin{subfigure}[b]{0.329\textwidth}
		\centering
		\includegraphics[width=\textwidth]{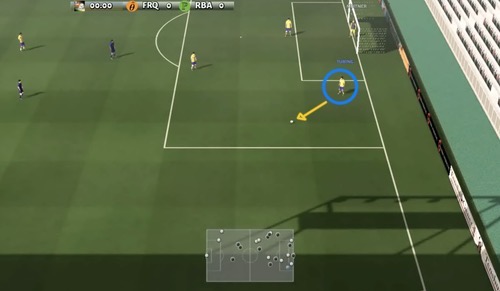}
		\caption{Initial player and ball positions in the level.~\vspace{.35cm}}
	\end{subfigure}
	\hfill
	\begin{subfigure}[b]{0.329\textwidth}
		\centering
		\includegraphics[width=\textwidth]{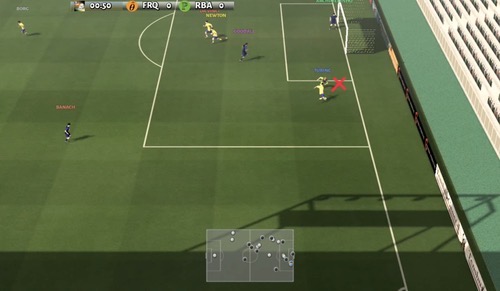}
		\caption{TiZero policy shoots from a narrow angle and is blocked by the goalkeeper}
	\end{subfigure}
	\hfill
	\begin{subfigure}[b]{0.329\textwidth}
		\centering
		\includegraphics[width=\textwidth]{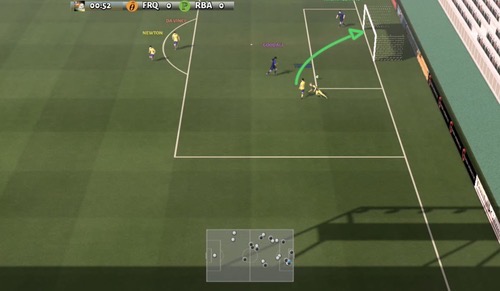}
		\caption{Reference policy goes to shoot from a better position and scores}
	\end{subfigure}
	\caption{Adversarial example of better shooting positioning.}
	\label{fig:shooting_position}
\end{figure*}

\begin{figure*}
	\begin{subfigure}[b]{0.329\textwidth}
		\centering
		\includegraphics[width=\textwidth]{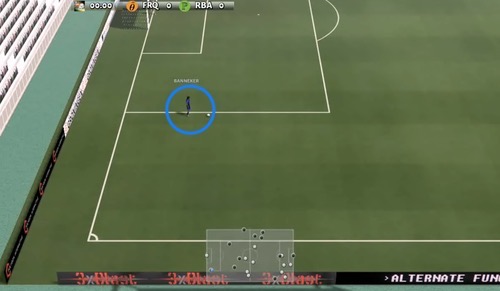}
		\caption{Initial player and ball positions in the level.~\vspace{.35cm}}
	\end{subfigure}
	\hfill
	\begin{subfigure}[b]{0.329\textwidth}
		\centering
		\includegraphics[width=\textwidth]{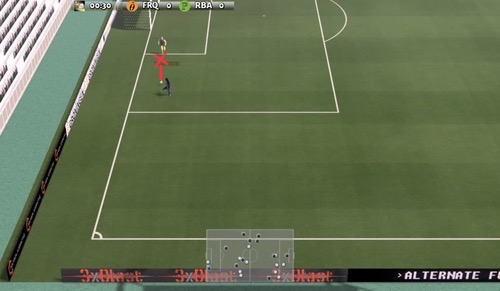}
		\caption{TiZero policy runs towards the goal and shoots, getting blocked by the goalkeeper.}
	\end{subfigure}
	\hfill
	\begin{subfigure}[b]{0.329\textwidth}
		\centering
		\includegraphics[width=\textwidth]{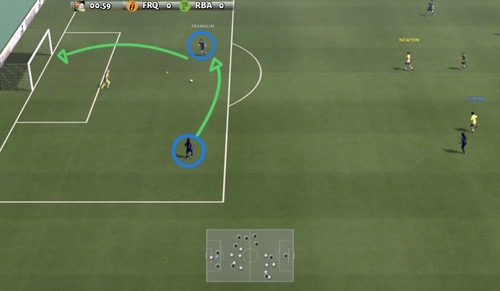}
		\caption{Reference policy passes the ball to a better-positioned player who scores.}
	\end{subfigure}
	\caption{Adversarial example of passing.} 
	\label{fig:passing}
\end{figure*}

\paragraph{\textbf{Unforced Own Goals}} Perhaps the most glaring adversarial behaviour discovered are instances where TiZero agents inexplicably shoot towards their own goal, resulting in unforced own goals~(See \cref{fig:owngoal}). In contrast, when starting from identical in-game positions, the reference policies manage to counterattack effectively, often resulting in successful scoring endeavours.

\paragraph{\textbf{Slow-running opponents}} The TiZero agents always choose to sprint throughout the episode. However, this makes them weak on defence against opponents who move slower with the ball. Instead of trying to tackle and take the ball, TiZero's main defensive strategy is to try and block opponents. Opponents can take advantage of this by using deceptive moves, especially when moving slowly, making it hard for TiZero's defenders to stop them. This is illustrated in \cref{fig:slow_running}.

\paragraph{\textbf{Suboptimal Ball Positioning for Shooting}} When trying to score a goal, TiZero agents often choose a suboptimal positioning, such as shooting from a narrow angle. In contrast, the reference policies often make subtle adjustments to optimally position the ball before initiating a shot (e.g., move towards the centre of the goals \cref{fig:shooting_position}). 

\paragraph{\textbf{Passing to Better Positioned Players}} 
A notable shortcoming in TiZero's policy, when compared to the built-in heuristic, is its reluctance to pass the ball to teammates who are in more favourable positions and have a higher likelihood of scoring, as illustrated in \cref{fig:passing}. In contrast, heuristic bots—whether easy, medium, or hard—demonstrate a consistent pattern of passing to optimally positioned players, enhancing their goal-scoring opportunities. This effective passing strategy seems unfamiliar to TiZero, causing it difficulty in overcoming a successful defence.

\section{Related Work}\label{sec:related_work}

\subsubsection*{\textbf{Quality Diversity}}
Quality Diversity (QD) is a category of open-ended learning methods aimed at discovering a collection of solutions that are both highly diverse and performant~\citep{lehman2011abandoning, Cully2018Quality}. 
Two commonly used QD algorithms are Novelty Search with Local Competition \citep[NSLC,][]{lehman2011abandoning} and MAP-Elites \citep{mouret2015illuminating, Cully2015RobotsTC}. These two approaches differ in the way they structure the archive; novelty search completely forgoes a grid and opts instead for growing an unstructured archive that dynamically expands, while MAP-Elites adopts a static mapping approach. Although \method{} leverages MAP-Elites as its diversity mechanism, it can be adapted to use NSLC. One of the most effective versions of MAP-Elites is CMA-ME \citep{fontaine2020coveriance}.
CMA-ME combines MAP-Elites with the evolutionary optimisation algorithm Covariance Matrix Adaptation Evolution Strategy (CMA-ES) \citep{cma-es}, improving the selection of the fittest solutions which will be perturbed to generate new elites. Mix-ME~\citep{ingvarsson2023mixme} extends MAP-Elites to multi-agent domains but is limited to fully cooperative settings.

\subsubsection*{\textbf{Multi-Agent RL}}
Recent advancements in the field of cooperative multi-agent RL~\citep{foerster2018counterfactual, rashid2018qmix, dewitt2020independent, mahajan2019maven} have shown remarkable success in tackling complex challenges in video games, such as StarCraft II~\citep{samvelyan2019starcraft, ellis2023smacv2}.
Google Research Football \citep[GRF,][]{kurach2020google} stands as one of the most complex multi-agent RL benchmarks, as a two-team zero-sum game with sparse reward and requiring a significant amount of coordination between co-players. Most of the prior work on addressing the toy settings of the GRF only involved a few agents (e.g., 3 vs 1 scenario).
Multi-Agent PPO \citep[MAPPO,][]{yu2022the} uses PPO \citep{schulman2017proximal} with a centralised critic to play on toy settings. %
CDS~\citep{li2021celebrating} analyses the importance of diversity between policies in GRF.
Multi-Agent Transformer~\citep[MAT,][]{wen2022multi} models GRF as a sequence problem using the self-attention mechanism. TiKick \citep{huang2021tikick} attempts to solve the full 11 vs 11 game using demonstrations from single-agent trajectories. SPC~\citep{wang2023towards} uses an adaptive curriculum on handcrafted environments for overcoming the sparse reward issue in GRF. TiZero is the first method that claims to have mastered the full 11 vs 11 game of GRF from scratch \citep{tizero} following $45$ days of training with a large amount of computational resources. To achieve this, TiZero uses a hand-crafted curriculum over environment variations, self-play, augmented observation space, reward shaping, and action masking. Of notable importance are also the works tackling the game of football in a robotics setting \citep{kitano1997robocup, kitano1997robocup2, riedmiller2009reinforcement}.%

\subsubsection*{\textbf{Adversarial attacks on multi-agent policies}} 
Deep neural networks, such as image classifiers, are known to be sensitive to adversarial attacks \citep{szegedy2013intriguing, carlini2019evaluating, ren2020adversarial}.
Such susceptibility has also been demonstrated in multi-agent RL.
\citet{wang2023adversarial} attacks the leading Go-playing AI, KataGo \citep{wu2019accelerating}, by training adversarial policies and achieving >97\% win rate against it. 
Such adversarial agents are not expert Go-playing bots and are easily defeated by amateur human players. Instead, they simply trick KataGo into making serious blunders. \citet{timbers2022approximate} introduce ISMCTS-BR, a search-based deep RL algorithm that learns a best response to a given agent.
Both of these solutions find exploitability using RL and expensive Monte-Carlo tree search \citep{mcts}, whereas \method{} is a fast, gradient-free, black-box method that finds adversarial settings using QD. Unlike the previous methods, \method{} is not restricted to any concrete agent architecture and is more general in nature.
MAESTRO~\citep{samvelyan2023maestro} crafts adversarial curricula for training robust agents in 2-player settings by jointly sampling environment/co-player pairs, emphasizing the interplay between agents and environments.

\section{Conclusion and Future Work}

This paper introduced \methodlong{} (\method{}), a novel approach aimed at systematically uncovering situations where pre-trained multi-agent RL agents display strategic errors. \method{} leverages quality-diversity mechanisms and employs regret to identify and quantify a multitude of scenarios where agents enact suboptimal strategies, with a particular focus on the advanced TiZero agent within the Google Research Football environment. Our investigations using \method{} revealed several previously unnoticed vulnerabilities in TiZero’s strategic decision-making, such as ineffective finishing and misunderstandings of the offside rule, highlighting the hidden strategic inefficiencies and latent vulnerabilities in even the most advanced RL agents.

Looking forward, we are eager to use \method{} in broader multi-agent domains, combining it with advanced evolutionary and learning strategies to enhance its ability to pinpoint limitations of multi-agent policies. Investigating various adversarial situations in pre-trained models will offer a deeper understanding of strategic complexities within multi-agent systems. Future research will also aim at using the discovered vulnerabilities in pre-trained models for enhancing the target model further through additional fine-tuning, thereby improving its robustness to unseen or adversarial settings.

\vfill\eject
\bibliographystyle{ACM-Reference-Format}
\balance
\bibliography{bib/refs}

\clearpage
\appendix
\section*{Acknowledgements}

We are grateful to Shiyu Huang for enabling our experiments with the TiZero agent and providing access to the checkpoints of TiZero saved throughout the training.
This work was funded by Meta.

\section{Adversarial Examples for Google Research Football}\label{appendix:football}

Below are 11 adversarial examples in TiZero we identify using \method{}.

\begin{figure*}
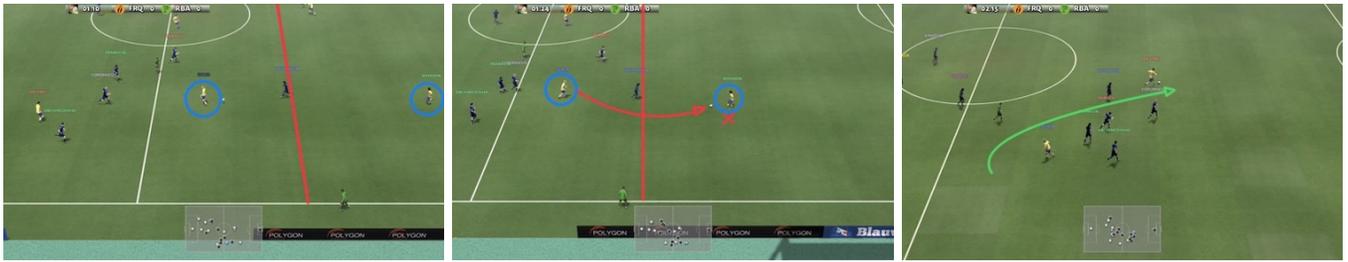

	\begin{subfigure}[b]{0.329\textwidth}
		\centering
		\includegraphics[width=\textwidth]{figures/cropped_examples/offside_a.jpg_cropped.jpg}
		\caption{Initial player and ball positions in the level. TiZero is about to pass the ball to a teammate.~\vspace{.35cm}} \end{subfigure}
	\hfill
	\begin{subfigure}[b]{0.329\textwidth}
		\centering
		\includegraphics[width=\textwidth]{figures/cropped_examples/offside_b.jpg_cropped.jpg}
		\caption{The receiving player is clearly offside, thus a freekick is awarded to the opponent's team.~\vspace{.35cm}}
	\end{subfigure}
	\hfill
	\begin{subfigure}[b]{0.329\textwidth}
		\centering
		\includegraphics[width=\textwidth]{figures/cropped_examples/offside_c.jpg_cropped.jpg}
		\caption{Reference policy does not pass to the offside player and directly runs towards the goal to score.}
	\end{subfigure}
	\caption{Adversarial example of offsides.}
	\label{fig:offside_A}
\end{figure*}

\paragraph*{\textbf{Offsides}} Despite its strong performance under standard evaluations, TiZero frequently falls victim to erroneously passing the ball to players unmistakably in offside positions, as shown in \cref{fig:offside_A}
This observation highlights TiZero's lack of a deep understanding of the rules of the game. In contrast, the reference policies abstain from passing the ball to offside players, resulting in successful scoring outcomes.\footnote{A player is offside when it is in the opponent's half and any part of their body is closer to the opponent's goal line than both the ball and the second-last opponent. Usually one of the two opponents is the goalkeeper. When this happens a free kick is awarded to the opponent's team.}

\begin{figure*}
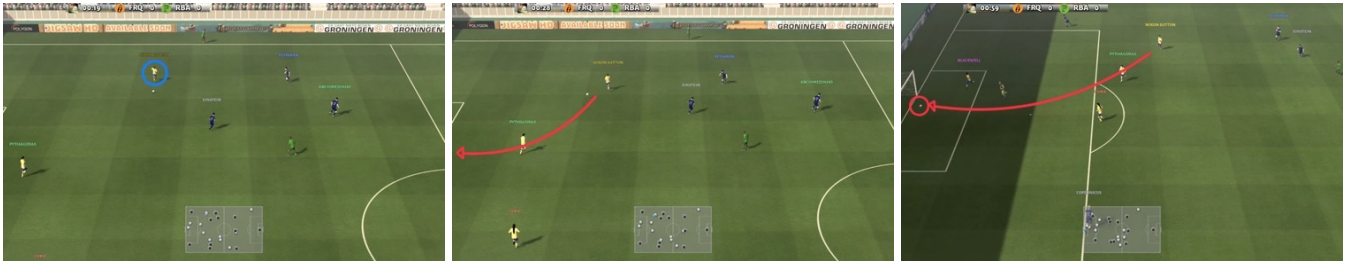

	\begin{subfigure}[b]{0.329\textwidth}
		\centering
		\includegraphics[width=\textwidth]{figures/cropped_examples/own_goal_a.jpg_cropped.jpg}
	\end{subfigure}
	\hfill
	\begin{subfigure}[b]{0.329\textwidth}
		\centering
		\includegraphics[width=\textwidth]{figures/cropped_examples/own_goal_b.jpg_cropped.jpg}
	\end{subfigure}
	\hfill
	\begin{subfigure}[b]{0.329\textwidth}
		\centering
		\includegraphics[width=\textwidth]{figures/cropped_examples/own_goal_c.jpg_cropped.jpg}
	\end{subfigure}
	\caption{Adversarial example of an own goal. TiZero gets tricked and shoots in its own goal.}
	\label{fig:owngoal_A}
\end{figure*}

\paragraph*{\textbf{Unforced Own Goals}} Perhaps the most glaring adversarial behaviour discovered are instances where TiZero agents inexplicably shoot towards their own goal, resulting in unforced own goals~(See \cref{fig:owngoal_A}). In contrast, when starting from identical in-game positions, the reference policies manage to counterattack effectively, often resulting in successful scoring endeavours.

\begin{figure*}
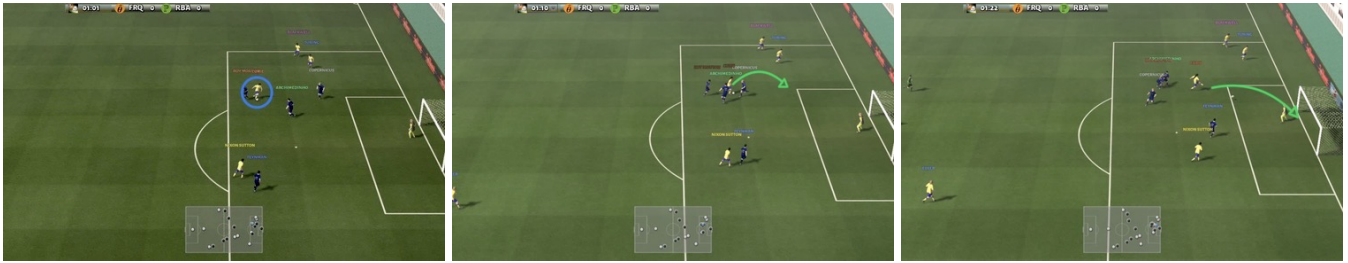

	\begin{subfigure}[b]{0.329\textwidth}
		\centering
		\includegraphics[width=\textwidth]{figures/cropped_examples/walking_1.jpg_cropped.jpg}
	\end{subfigure}
	\hfill
	\begin{subfigure}[b]{0.329\textwidth}
		\centering
		\includegraphics[width=\textwidth]{figures/cropped_examples/walking_2.jpg_cropped.jpg}
	\end{subfigure}
	\hfill
	\begin{subfigure}[b]{0.329\textwidth}
		\centering
		\includegraphics[width=\textwidth]{figures/cropped_examples/walking_3.jpg_cropped.jpg}
	\end{subfigure}
	\caption{Adversarial example of a slow-running opponent. Three TiZero-controlled defenders are not able to stop a simple slow-running opponent controlled by the reference policy, who walks past them and scores.}
	\label{fig:slow_running_A}
\end{figure*}

\paragraph*{\textbf{Slow-running opponents}} The TiZero agents always choose to sprint throughout the episode. However, this makes them weak in defence against opponents who move slower with the ball. Instead of trying to tackle and take the ball, TiZero's main defensive strategy is to try and block opponents. Opponents can take advantage of this by using deceptive moves, especially when moving slowly, making it hard for TiZero's defenders to stop them. This is illustrated in \cref{fig:slow_running_A}.

\begin{figure*}
	\begin{subfigure}[b]{0.329\textwidth}
		\centering
		\includegraphics[width=\textwidth]{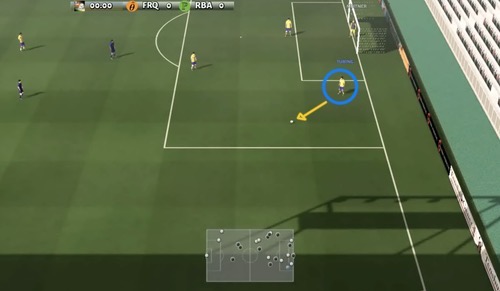}
		\caption{Initial player and ball positions in the level.~\vspace{.35cm}}
	\end{subfigure}
	\hfill
	\begin{subfigure}[b]{0.329\textwidth}
		\centering
		\includegraphics[width=\textwidth]{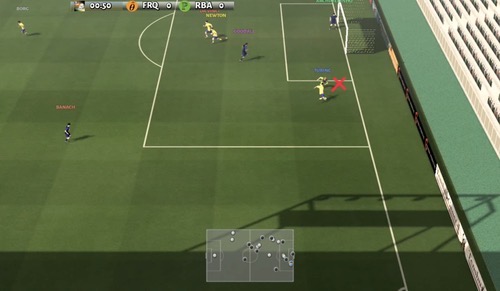}
		\caption{TiZero shoots from a narrow angle and is blocked by the goalkeeper}
	\end{subfigure}
	\hfill
	\begin{subfigure}[b]{0.329\textwidth}
		\centering
		\includegraphics[width=\textwidth]{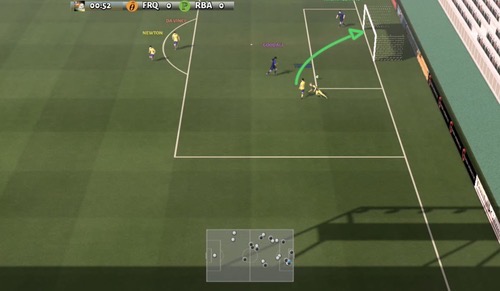}
		\caption{Reference policy goes to shoot from a better position and scores}
	\end{subfigure}
	\caption{Adversarial example of better shooting positioning.}
	\label{fig:shooting_position_A}
\end{figure*}

\paragraph*{\textbf{Suboptimal Ball Positioning for Shooting}} When trying to score a goal, TiZero agents often choose a suboptimal positioning, such as shooting from a narrow angle. In contrast, the reference policies often make subtle adjustments to optimally position the ball before initiating a shot (e.g., move towards the centre of the goals \cref{fig:shooting_position_A}). 

\begin{figure*}
	\begin{subfigure}[b]{0.329\textwidth}
		\centering
		\includegraphics[width=\textwidth]{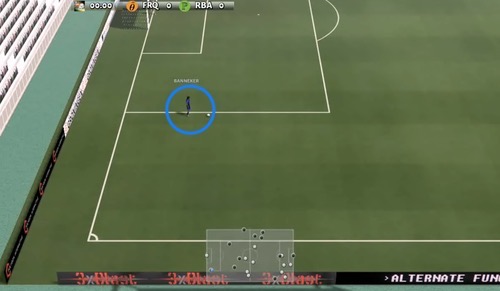}
		\caption{Initial player and ball positions in the level.~\vspace{.35cm}}
	\end{subfigure}
	\hfill
	\begin{subfigure}[b]{0.329\textwidth}
		\centering
		\includegraphics[width=\textwidth]{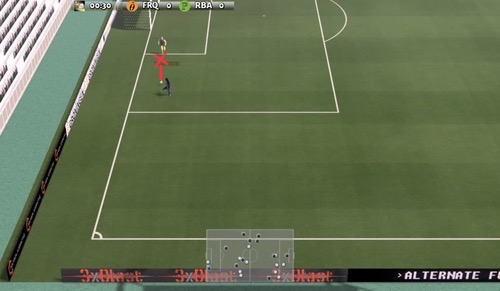}
		\caption{TiZero runs towards the goal and shoots, getting blocked by the goalkeeper.}
	\end{subfigure}
	\hfill
	\begin{subfigure}[b]{0.329\textwidth}
		\centering
		\includegraphics[width=\textwidth]{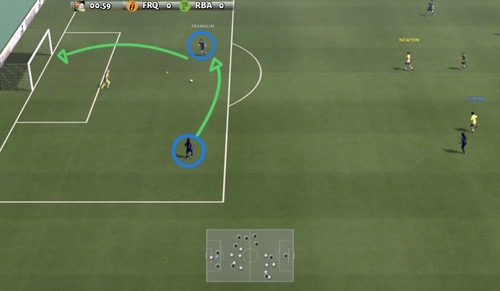}
		\caption{Reference policy passes the ball to a better-positioned player who scores.}
	\end{subfigure}
	\caption{Adversarial example of passing.} 
	\label{fig:passing_A}
\end{figure*}

\paragraph*{\textbf{Passing to Better Positioned Players}} 
A notable shortcoming in TiZero's policy, when compared to the built-in heuristic, is its reluctance to pass the ball to teammates who are in more favourable positions and have a higher likelihood of scoring, as illustrated in \cref{fig:passing_A}. In contrast, heuristic bots—whether easy, medium, or hard—demonstrate a consistent pattern of passing to optimally positioned players, enhancing their goal-scoring opportunities. This effective passing strategy seems unfamiliar to TiZero, causing it difficulty in overcoming a successful defence.

\begin{figure*}
	\begin{subfigure}[b]{0.329\textwidth}
		\centering
		\includegraphics[width=\textwidth]{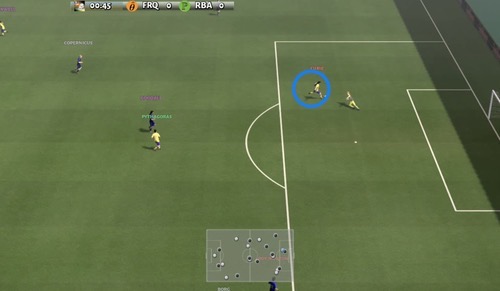}
		\caption{Initial player and ball positions in the level.~\vspace{.35cm}}
	\end{subfigure}
	\hfill
	\begin{subfigure}[b]{0.329\textwidth}
		\centering
		\includegraphics[width=\textwidth]{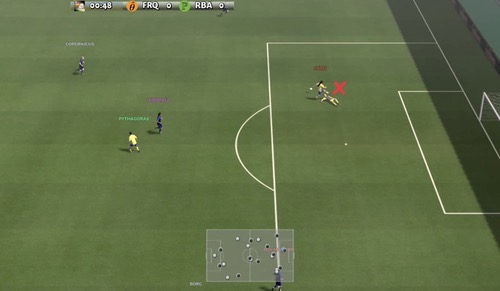}
		\caption{TiZero shoots while sprinting and the ball gets blocked by the goalkeeper.}
	\end{subfigure}
	\hfill
	\begin{subfigure}[b]{0.329\textwidth}
		\centering
		\includegraphics[width=\textwidth]{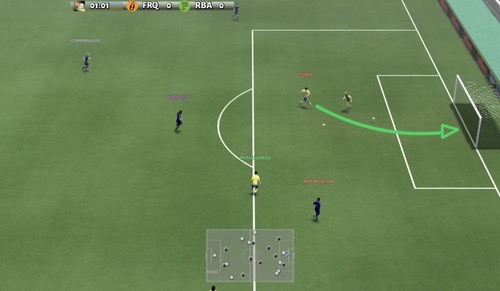}
		\caption{Reference policy doesn't run and is able to score.}
	\end{subfigure}
	\caption{Adversarial example of shooting while running.}
	\label{fig:running_shooting_A}
\end{figure*}

\paragraph*{\textbf{Shooting while Running}}
Capitalizing on other game mechanics, the reference policies exhibit stronger behaviours by halting their sprinting behaviour leading up to a shot, resulting in a notably higher success rate in goal realisation. TiZero's agents, in contrast, consistently maintain a sprinting stance, thereby frequently missing straightforward scoring opportunities in front of the opposing goalkeepers (\cref{fig:running_shooting_A}).

\begin{figure*}
	\begin{subfigure}[b]{0.329\textwidth}
		\centering
		\includegraphics[width=\textwidth]{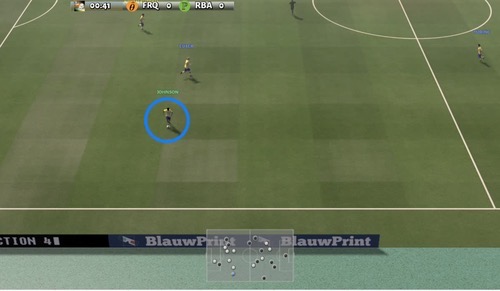}
		\caption{Initial player and ball positions in the level.~\vspace{.35cm}}
	\end{subfigure}
	\hfill
	\begin{subfigure}[b]{0.329\textwidth}
		\centering
		\includegraphics[width=\textwidth]{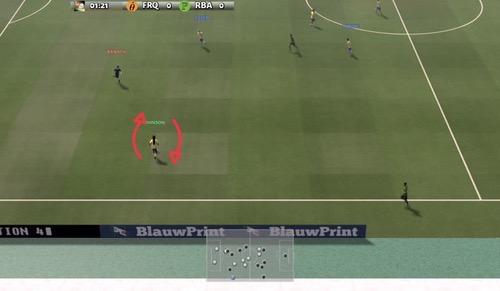}
		\caption{TiZero aimlessly runs up and down from the same position in an endless loop.}
	\end{subfigure}
	\hfill
	\begin{subfigure}[b]{0.329\textwidth}
		\centering
		\includegraphics[width=\textwidth]{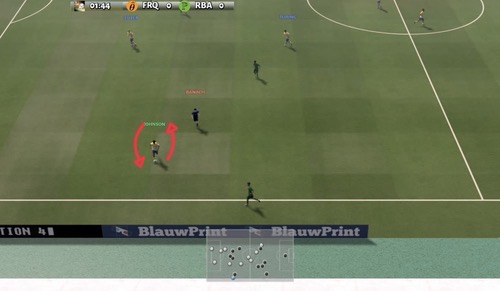}
		\caption{Reference policy attacks the opponent's goal, often resulting in goal-scoring endeavours.}
	\end{subfigure}
	\caption{Adversarial example of confused behaviour.}
	\label{fig:confused_A}
\end{figure*}

\paragraph*{\textbf{Confused Agent Behaviour}}
Another intriguing adversarial instance finds TiZero's ball-possessing player aimlessly sprinting back and forth in random areas of the field, thereby exhibiting a completely unproductive pattern of movement (\cref{fig:confused_A}).

\begin{figure*}
	\begin{subfigure}[b]{0.329\textwidth}
		\centering
		\includegraphics[width=\textwidth]{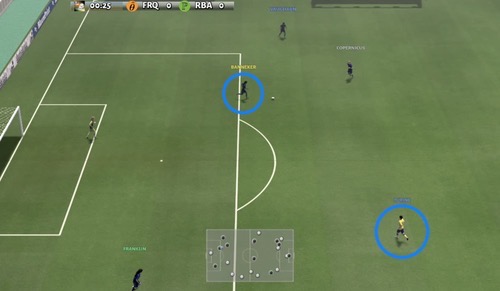}
		\caption{Initial player and ball positions in the level.~\vspace{.7cm}}
	\end{subfigure}
	\hfill
	\begin{subfigure}[b]{0.329\textwidth}
		\centering
		\includegraphics[width=\textwidth]{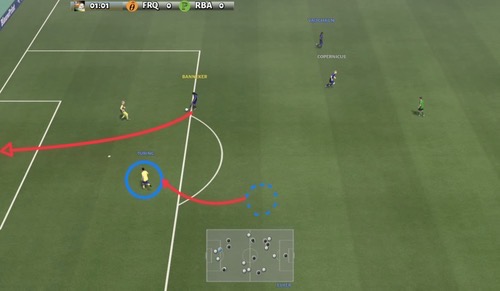}
		\caption{TiZero's defender runs along a suboptimal trajectory, giving space for the opponent to shoot and score.}
	\end{subfigure}
	\hfill
	\begin{subfigure}[b]{0.329\textwidth}
		\centering
		\includegraphics[width=\textwidth]{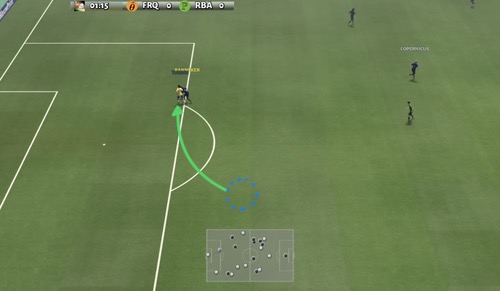}
		\caption{Reference policy instead runs towards the attacker to block the attempt.~\vspace{.35cm}}
	\end{subfigure}
	\caption{Adversarial example of better defensive behaviour.}
	\label{fig:defense_A}
\end{figure*}

\paragraph*{\textbf{Improved Defensive Positioning}}
TiZero shows several vulnerabilities in its defensive strategies, failing to close down on the opponent's attacking trajectory and allowing them to score. In comparison, \cref{fig:defense_A} shows the reference policies closing down on the opponent striker and seizing the ball before they have the chance to shoot.

\begin{figure*}
	\begin{subfigure}[b]{0.329\textwidth}
		\centering
		\includegraphics[width=\textwidth]{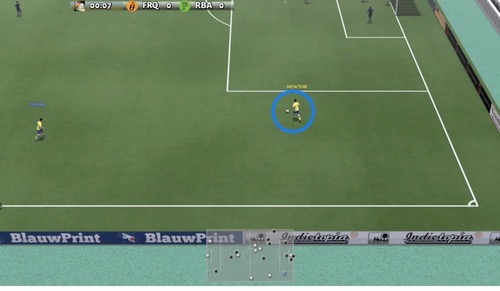}
		\caption{Initial player and ball positions in the level.~\vspace{.35cm}}
	\end{subfigure}
	\hfill
	\begin{subfigure}[b]{0.329\textwidth}
		\centering
		\includegraphics[width=\textwidth]{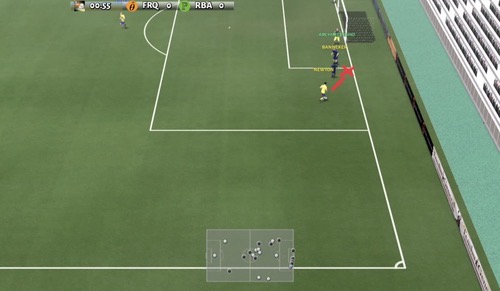}
		\caption{TiZero's team runs backwards, leaving a solitary attacker confused and unable to score.}
	\end{subfigure}
	\hfill
	\begin{subfigure}[b]{0.329\textwidth}
		\centering
		\includegraphics[width=\textwidth]{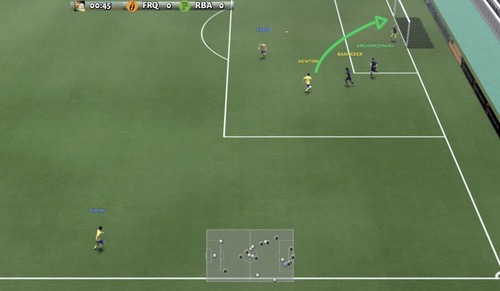}
		\caption{Reference policy instead doesn't get tricked, the attacker moves in a better position to score.}
	\end{subfigure}
	\caption{Adversarial example of erroneous team movement.}
	\label{fig:team_movement_A}
\end{figure*}

\paragraph*{\textbf{Erroneous Team Movement}}
Several adversarial examples show the entirety of TiZero's team running in the wrong direction to defend their goal, while the ball is positioned favourably towards the opponent's goal, leaving a solitary attacking player without support, who gets deceived and performs poorly. The reference policy instead doesn't get tricked and often manages to score despite the disarray (\cref{fig:team_movement_A}).

\begin{figure*}
	\begin{subfigure}[b]{0.329\textwidth}
		\centering
		\includegraphics[width=\textwidth]{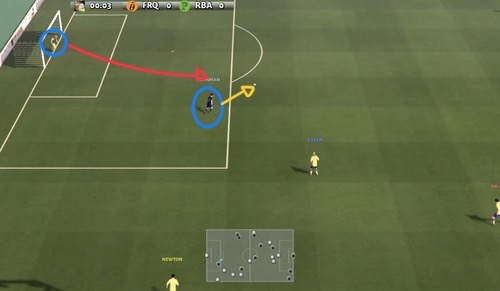}
		\caption{Initial player and ball positions in the level.~\vspace{.7cm}}
	\end{subfigure}
	\hfill
	\begin{subfigure}[b]{0.329\textwidth}
		\centering
		\includegraphics[width=\textwidth]{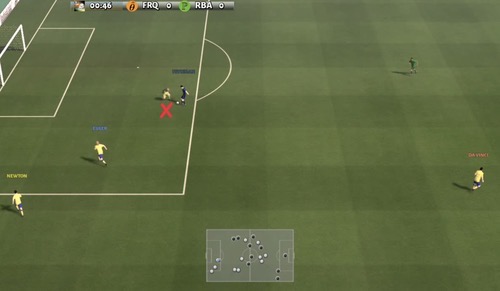}
		\caption{TiZero hesitates before shooting, giving enough time for the goalkeeper to seize the ball}
	\end{subfigure}
	\hfill
	\begin{subfigure}[b]{0.329\textwidth}
		\centering
		\includegraphics[width=\textwidth]{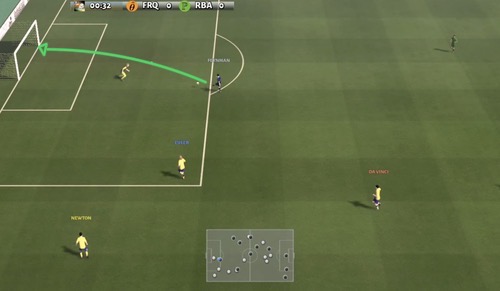}
		\caption{Reference policy instead shoots without hesitation and scores.~\vspace{.35cm}}
	\end{subfigure}
	\caption{Adversarial example of hesitation before shooting.}
	\label{fig:hesitation_A}
\end{figure*}

\paragraph*{\textbf{Hesitation Before Shooting}} The most common adversarial scenario encountered by the heuristic bots is situations in which TiZero hesitates before taking a shot, allowing the goalkeeper or defending players to seize the ball. In contrast, the inbuilt bot promptly recognizes the opportunity and shoots without hesitation, resulting in successful scoring (\cref{fig:hesitation_A}).

\begin{figure*}
	\begin{subfigure}[b]{0.329\textwidth}
		\centering
		\includegraphics[width=\textwidth]{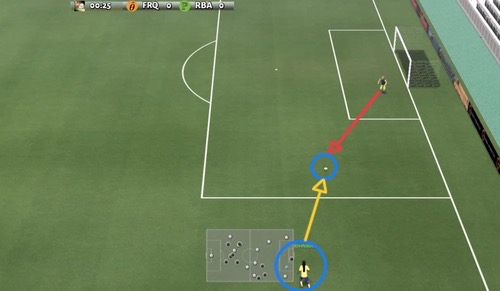}
		\caption{Initial player and ball positions in the level.~\vspace{.7cm}}
	\end{subfigure}
	\hfill
	\begin{subfigure}[b]{0.329\textwidth}
		\centering
		\includegraphics[width=\textwidth]{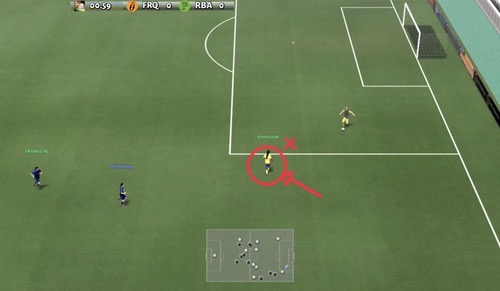}
		\caption{TiZero's attacker does not realise it can get to the ball before the goalkeeper and runs backwards.}
	\end{subfigure}
	\hfill
	\begin{subfigure}[b]{0.329\textwidth}
		\centering
		\includegraphics[width=\textwidth]{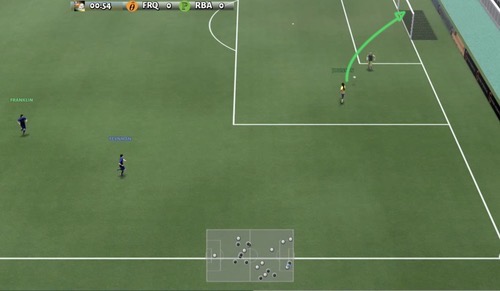}
		\caption{Reference policy instead runs towards the ball, reaching it before the goalkeeper does and scoring.}
	\end{subfigure}
	\caption{Adversarial example of missing a goal-scoring opportunity.}
	\label{fig:realizing_A}
\end{figure*}

\paragraph*{\textbf{Missing a Goal Scoring Opportunity}} TiZero often fails to acknowledge easy goal-scoring opportunities, where it could get to the ball and score, but instead decides not to pursue it. \cref{fig:realizing_A} shows how the reference policy capitalises on this kind of opportunity and scores.

\section{Environment Details}\label{appendix:env}

In our experiments with Google Research Football~\cite{kurach2020google}, we adopt a procedural generation method for level creation. For each player, as well as the ball, we randomly sample the $(x,y)$ coordinates: the x-coordinate is sampled from the range $[-0.9, 0.9]$ and the y-coordinate from the range $[-0.4, 0.4]$. The settings employed during the generation are as follows:

\begin{itemize}
	\item \texttt{deterministic}: set to \texttt{False}, implying that levels can have non-deterministic components.
	\item \texttt{offsides}: set to \texttt{True}, enforcing the offsides rule during gameplay.
	\item \texttt{end\_episode\_on\_score}: set to \texttt{True}, which means the episode will terminate once a goal is scored.
	\item \texttt{end\_episode\_on\_out\_of\_play}: set to \texttt{False}, indicating the episode will not end on ball out-of-play events.
	\item \texttt{end\_episode\_on\_possession\_change}: set to \texttt{False}, indicating the episode will not end when the ball changes possession from one team to another.
\end{itemize}

For the \textit{easy} bot, the difficulty is set at $0.05$. For the \textit{medium} bot, it is set to $0.5$, and for the \textit{hard} bot, the difficulty is at $0.95$. These values serve as the defaults in GRF, ensuring consistency across different game scenarios

We use the enhanced observation space as described in TiZero~\citep{tizero}, consisting of a $268$-dimensional vector including information.

\section{Implementation Details}\label{appendix:implementation}

Hyperparameters of \method{} are provided in \cref{table:hyperparams}. We use the CMA-ME as implemented in pyribs~\citep{pyribs}. 
For the TiZero and reference agents, we use the exact agent architecture as in the original paper~\citep{tizero} using TiZero's official open-source release~\citep{tizero_github}. Parameter sharing is applied to all agents in the team.

\begin{table}[t!]
	\caption{Hyperparameters used for finding adversarial examples in Google Research Football.}
	\label{table:hyperparams}
	\begin{center}
		\scalebox{0.87}{
			\begin{tabular}{lr}
				\toprule
				\textbf{Parameter} &  \\
				\midrule
				Number of steps & 5000 \\
				Game duration & 128 \\
				Number of CMA-ME emitters & 4 \\
				Number of repeats per level & 4 \\
				Emitter gaussian noise $\sigma$ & 0.1 \\
				Ranker & improvement \\
				QD score offset & -2 \\

				\bottomrule 
			\end{tabular}
		}
	\end{center}
\end{table}

The policy network is made up of six different multi-layer perceptrons (MLPs), each having two fully connected layers, including one specifically for the 'player ID', to encode every part of the observation individually. The MLP layers have a hidden size of 64. The hidden features extracted are brought together and then handled by an LSTM layer to give the agent memory, with the hidden size for this layer being 256. Every hidden layer is equipped with layer normalisation and ReLU non-linearities. The orthogonal matrix is used for initialising parameters, and the learning process is optimized with the Adam optimizer. Similar to the original implementation, illegal actions are masked out by making their selection probability zero. The action output layer utilises a softmax layer and is formed with a 19-dimension vector. 

Experiments are conducted on an in-house cluster. Every task, denoted by a seed, uses one Tesla V100 GPU and 10 CPUs. For each of the $51$ reference policies ($48$ TiZero checkpoints and $3$ built-in bots), we use 3 random seeds, for each of the baselines. Runs last approximately $8.5$ days for $5000$ iterations of \method{}.

\end{document}